\documentclass{article}

\PassOptionsToPackage{numbers, compress}{natbib}

\usepackage{microtype}
\usepackage{graphicx}
\usepackage{subfigure}
\usepackage{booktabs} % for professional tables

\usepackage{hyperref}

\usepackage{wrapfig}
\usepackage{overpic}

\usepackage{amsmath,amsfonts,bm}

\def\eqref#1{equation~\ref{#1}}
\def\1{\bm{1}}

\DeclareMathAlphabet{\mathsfit}{\encodingdefault}{\sfdefault}{m}{sl}
\SetMathAlphabet{\mathsfit}{bold}{\encodingdefault}{\sfdefault}{bx}{n}

\usepackage{hyperref}
\usepackage{url}
\usepackage{xcolor}

\usepackage{graphicx}
\usepackage[english]{babel}
\usepackage{soul}% for \st

\usepackage[preprint]{neurips_2020}
\usepackage[utf8]{inputenc} % allow utf-8 input
\usepackage[T1]{fontenc}    % use 8-bit T1 fonts
\usepackage{hyperref}       % hyperlinks
\usepackage{url}            % simple URL typesetting
\usepackage{booktabs}       % professional-quality tables
\usepackage{amsfonts}       % blackboard math symbols
\usepackage{nicefrac}       % compact symbols for 1/2, etc.
\usepackage{microtype}      % microtypography

\definecolor{darkgreen}{rgb}{0,0.6,0}
\definecolor{darkred}{rgb}{0.7,0.0,0}
\definecolor{darkblue}{rgb}{0,0.0,0.6}
\definecolor{magenta}{rgb}{0.8,0.1,0.8}
\definecolor{darksomething}{rgb}{0,0.4,0.6}

\newcommand{\sref}[1]{\S\ref{#1}}

\usepackage{color}
\definecolor{deepblue}{rgb}{0,0,0.5}
\definecolor{deepred}{rgb}{0.6,0,0}
\definecolor{deepgreen}{rgb}{0,0.5,0}

\usepackage{listings}

\definecolor{gray}{gray}{0.5}
\colorlet{commentcolour}{green!50!black}

\colorlet{stringcolour}{red!60!black}
\colorlet{keywordcolour}{magenta!90!black}
\colorlet{exceptioncolour}{yellow!50!red}
\colorlet{commandcolour}{blue!60!black}
\colorlet{numpycolour}{blue!60!green}
\colorlet{literatecolour}{magenta!90!black}
\colorlet{promptcolour}{green!50!black}
\colorlet{specmethodcolour}{violet}

\newcommand*{\literatecolour}{\textcolor{literatecolour}}

\newcommand*{\pythonprompt}{\textcolor{promptcolour}{{>}{>}{>}}}

\lstdefinestyle{mypython}{
language=python,
showtabs=true,
tab=,
tabsize=2,
basicstyle=\ttfamily\footnotesize,%\setstretch{.5},
stringstyle=\color{stringcolour},
showstringspaces=false,
alsoletter={1234567890},
otherkeywords={\%, \}, \{, \&, \|},
keywordstyle=\color{keywordcolour}\bfseries,
emph={and,break,class,continue,def,yield,del,elif ,else,%
except,exec,finally,for,from,global,if,import,in,%
lambda,not,or,pass,print,raise,return,try,while,assert,with},
emphstyle=\color{blue}\bfseries,
emph={[2]True, False, None},
emphstyle=[2]\color{keywordcolour},
emph={[3]object,type,isinstance,copy,deepcopy,zip,enumerate,reversed,list,set,len,dict,tuple,xrange,append,execfile,real,imag,reduce,str,repr},
emphstyle=[3]\color{commandcolour},
emph={Exception,NameError,IndexError,SyntaxError,TypeError,ValueError,OverflowError,ZeroDivisionError},
emphstyle=\color{exceptioncolour}\bfseries,
morecomment=[s]{"""}{"""},
commentstyle=\color{commentcolour}\slshape,
emph={[4]ode, fsolve, sqrt, exp, sin, cos,arctan, arctan2, arccos, pi,  array, norm, solve, dot, arange, isscalar, max, sum, flatten, shape, reshape, find, any, all, abs, plot, linspace, legend, quad, polyval,polyfit, hstack, concatenate,vstack,column_stack,empty,zeros,ones,rand,vander,grid,pcolor,eig,eigs,eigvals,svd,qr,tan,det,logspace,roll,min,mean,cumsum,cumprod,diff,vectorize,lstsq,cla,eye,xlabel,ylabel,squeeze},
emphstyle=[4]\color{numpycolour},
emph={[5]__init__,__add__,__mul__,__div__,__sub__,__call__,__getitem__,__setitem__,__eq__,__ne__,__nonzero__,__rmul__,__radd__,__repr__,__str__,__get__,__truediv__,__pow__,__name__,__future__,__all__},
emphstyle=[5]\color{specmethodcolour},
emph={[6]assert,yield},
emphstyle=[6]\color{keywordcolour}\bfseries,
emph={[7]range},
emphstyle={[7]\color{keywordcolour}\bfseries},
literate=*%
{:}{{\literatecolour:}}{1}%
{=}{{\literatecolour=}}{1}%
{-}{{\literatecolour-}}{1}%
{+}{{\literatecolour+}}{1}%
{*}{{\literatecolour*}}{1}%
{**}{{\literatecolour{**}}}2%
{/}{{\literatecolour/}}{1}%
{//}{{\literatecolour{//}}}2%
{!}{{\literatecolour!}}{1}%
{[}{{\literatecolour[}}{1}%
{]}{{\literatecolour]}}{1}%
{<}{{\literatecolour<}}{1}%
{>}{{\literatecolour>}}{1}%
{>>>}{\pythonprompt}{3}%
,%
frame=trbl,
rulecolor=\color{black!40},
backgroundcolor=\color{white},
breakindent=.5\textwidth,frame=single,breaklines=true%
}

\lstnewenvironment{python}[1][]{\lstset{style=mypython}}{}

\lstdefinestyle{mypythoninline}{
style=mypython,%
basicstyle=\ttfamily,%
keywordstyle=\color{keywordcolour},%
emphstyle={[7]\color{keywordcolour}},%
emphstyle=\color{exceptioncolour},%
literate=*%
{:}{{\literatecolour:}}{2}%
{=}{{\literatecolour=}}{2}%
{-}{{\literatecolour-}}{2}%
{+}{{\literatecolour+}}{2}%
{*}{{\literatecolour*}}2%
{**}{{\literatecolour{**}}}3%
{/}{{\literatecolour/}}{2}%
{//}{{\literatecolour{//}}}{2}%
{!}{{\literatecolour!}}{2}%
{[}{{\literatecolour[}}{2}%
{]}{{\literatecolour]}}{2}%
{<}{{\literatecolour<}}{2}%
{<=}{{\literatecolour{<=}}}3%
{>}{{\literatecolour>}}{2}%
{>=}{{\literatecolour{>=}}}3%
{==}{{\literatecolour{==}}}3%
{!=}{{\literatecolour{!=}}}3%
{+=}{{\literatecolour{+=}}}3%
{-=}{{\literatecolour{-=}}}3%
{*=}{{\literatecolour{*=}}}3%
{/=}{{\literatecolour{/=}}}3%
}

\title{Tasks, stability, architecture, and compute:\\
Training more effective learned optimizers,\\ and using them to train themselves}

\author{%
  Luke Metz \\
  Google Research, Brain Team \\
  \texttt{lmetz@google.com} \\
  \And
  Niru Maheswaranathan \\ 
    Google Research, Brain Team \\
  \texttt{nirum@google.com} \\
  \AND
  C. Daniel Freeman \\
  Google Research, Brain Team \\
  \texttt{cdfreeman@google.com}\\
  \And
  Ben Poole \\
  Google Research, Brain Team \\
  \texttt{pooleb@google.com}  \\
  \And
  Jascha Sohl-Dickstein \\
  Google Research, Brain Team \\
  \texttt{jaschasd@google.com}  
}

\begin{document}

\maketitle

\begin{abstract}
Much as replacing hand-designed features with learned functions has revolutionized how we solve perceptual tasks, we believe learned algorithms will transform how we train models.
In this work we focus on general-purpose learned optimizers capable of training a wide variety of problems with no user-specified hyperparameters.
We introduce a new, neural network parameterized, hierarchical optimizer with access to additional features such as validation loss to enable automatic regularization. 
Most learned optimizers have been trained on only a single task, or a small number of tasks.
We train our optimizers on thousands of tasks, making use of orders of magnitude more compute, resulting in optimizers that generalize better to unseen tasks.
The learned optimizers not only perform well, but learn behaviors that are distinct from existing first order optimizers. 
For instance, they generate update steps that have implicit regularization and adapt as the problem hyperparameters (e.g. batch size) or architecture (e.g. neural network width) change.
Finally, these learned optimizers show evidence of being useful for out of distribution tasks such as training themselves from scratch.
\end{abstract}

\section{Introduction}
Much of the success of modern deep learning has been driven by a shift from hand-designed features carefully curated by human experts, to domain-agnostic methods that can learn features from large amounts of data. By leveraging large-scale datasets with flexible models, we are now able to rapidly learn powerful features for new problem settings that often generalize to novel tasks.
While learned features outperform hand-designed features on numerous tasks \citep{krizhevsky2012imagenet, berner2019dota, vinyals2019alphastar, piech2015deep}, we continue to use hand-designed optimization algorithms (such as gradient descent, momentum, and so on) for training models.

These hand-designed update rules benefit from decades of optimization research but still require extensive expert supervision in order to be used effectively in machine learning. For example, they fail to flexibly adapt to new problem settings and require careful tuning of learning rate schedules and momentum timescales for different model architectures and datasets \citep{choi2019empirical}. In addition, most do not leverage alternative sources of information beyond the gradient, such as the validation loss.
By separating the optimization target (training loss) from the broader goal (generalization), classic methods require more careful tuning of regularization and/or data augmentation strategies by the practitioner.

To address these drawbacks, recent work on \textit{learned} optimizers aims to replace hand-designed optimizers with a parametric optimizer, trained on a set of tasks, that can then be applied more generally. Recent work in this area has focused on either augmenting existing optimizers to adapt their own hyperparameters \citep{daniel2016learning, xu2017reinforcement, xu2019learning}, or developing more expressive learned optimizers to replace existing optimizers entirely \citep{andrychowicz2016learning, wichrowska2017learned, lv2017learning, metz2018learning, metz2019using, metz2019understanding,gu2019meta}. These latter models take in problem information (such as the current gradient of the training loss) and iteratively update parameters. However, to date, learned optimizers have proven to be brittle and ineffective at generalizing across diverse sets of problems.

Our work identifies fundamental barriers that have limited progress in learned optimizer research and addresses several of these barriers to train effective optimizers: 
\begin{enumerate}
    \item {\bf Computational scale}: Training a learned optimizer is costly. When training the optimizer, a single training step requires applying the optimizer to a training task for some number of unrolled steps. This work utilizes massive parallel computing infrastructure to scale the number of unrolled steps an order of magnitude larger than in previous work.
    \item {\bf Training tasks}: Deep learning requires large training datasets. For learned optimizers to be effective, we similarly need a large dataset of \textit{optimization tasks} on which to train the optimizer. We build off of the TaskSet dataset~\citep{metz2020using} and construct a dataset of more than a thousand diverse optimization tasks commonly found in machine learning. We show how this large and diverse task distribution is critical for training optimizers that generalize.
    \item {\bf Inductive bias of optimizer architecture}: 
    The parameterization of the learned optimizer and the task information fed to it
    both strongly affect performance. In this work, we propose a new hierarchical learned optimizer architecture that incorporates additional task information (such as validation loss), and show that it outperforms previous learned optimizer architectures.
\end{enumerate}

By addressing these barriers, we develop learned optimizers that exceed prior work in scale, robustness, and out of distribution generalization.
As a final test, we show that the learned optimizer can be used to train new learned optimizers from scratch (analogous to ``self-hosting'' compilers \citep{hart1962ai}).
We see this final accomplishment as being analogous to the first time a compiler is complete enough that it can be used to compile itself. 

\section{Preliminaries}
\label{sec:preliminaries}
Training a learned optimizer is a bilevel optimization problem that contains two loops: an \textit{inner} loop that applies the optimizer to solve a task, and an \textit{outer} loop that iteratively updates the parameters of the learned optimizer~\citep{franceschi2018bilevel}. We use the \textit{inner-} and \textit{outer-} prefixes throughout to be explicit about which optimization loop we are referring to. That is, the \textit{inner-loss} refers to a target task's loss function that we wish to optimize, and the \textit{outer-loss} refers to a measure of the optimizer's performance training the target task (inner-task). Correspondingly, we refer to the optimizer parameters as \textit{outer-parameters}, and the parameters that the optimizer is updating as \textit{inner-parameters}.
\textit{Outer-optimization} refers to the act of finding outer-parameters that perform well under some \textit{outer-loss}.

For a given inner-task, we apply the learned optimizer for some number of steps (\textit{unrolling} the optimizer). Ideally, we would unroll each target task until some stopping criterion is reached, but this is computationally infeasible for even moderate scale machine learning tasks. \textit{Each} outer-iteration requires unrolling the optimizer on a target task. Short (truncated) unrolls are more computationally efficient, but suffer from truncation bias~\cite{wuunderstanding, metz2019understanding} in that the outer-loss surface computed using truncated unrolls is different (and may have different minima) than the fully unrolled outer-loss.

\section{Methods: Addressing the three barriers to learned optimizers}
\label{sec:methods}

\subsection{Outer-Optimization}
\label{sec:outer_opt}
To train the optimizer, we minimize an outer-loss that quantifies the performance of the optimizer. This is defined as the mean of the inner-loss computed on the inner-{\em{validation}} set for some number of unrolled steps, averaged over inner-tasks in the outer-training taskset.
Although this outer-loss is differentiable, it is costly to compute the outer-gradient (which involves backpropagating through the unrolled optimization).
In addition, the outer-loss surface is badly conditioned and extremely non-smooth~\cite{metz2019understanding}, making it difficult to optimize.

We deal with these issues by using derivative-free optimization--specifically, evolutionary strategies (ES)~\cite{rechenberg1973evolutionsstrategie}--to minimize the outer-loss, obviating the need to compute derivatives through the unrolled optimization process. Previous work has used unrolled derivatives  \citep{andrychowicz2016learning,wichrowska2017learned,metz2019understanding}, and was thus limited to short numbers of unrolled steps (e.g. 20 in \citet{andrychowicz2016learning} and starting at 50 in \citet{metz2019understanding}). Using ES, we are able to use considerably longer unrolls. Initial unroll lengths were chosen to balance communication cost between parallel workers (when updating optimizer parameters) with the computational cost of unrolling on individual workers (when estimating the local gradient with ES).
We start outer-training by sampling unroll steps uniformly from 240-360 steps. When performance saturates, we continue training with Persistent Evolotionary Strategies (PES)~\citep{pes}. PES provides an unbiased estimate of the outer-gradient over the entire inner task, but at the cost of higher variance gradients.

ES and PES have an additional benefit, in that optimizing with ES smooths the underlying loss function. This smoothing helps stabilize outer-training~\citep{metz2019understanding}.
We set the standard deviation of the Gaussian distribution used by the ES algorithm (which also controls how much the outer-loss is smoothed) to 0.01. To deal with the high variance of the ES estimate of the gradient, we use antithetic sampling and train in in parallel using 1024 multi-core CPU workers.
While using more workers increases training speed, we find 1024 to be the point where performance gains become sub-linear. For more details see Appendix \ref{app:outer_opt}.

\subsection{Task distributions} \label{sec:task_dist}
To train the optimizer, we need to define a set of inner-tasks to use for training.
The choice of training tasks is critically important for two reasons: it determines the ability of the optimizer to outer-generalize (i.e. the learned optimizer's performance on new tasks), and it determines the computational complexity of outer-training.
For improved outer-generalization, we would like our inner-problems to be representative of tasks we care about. In this work, these are state-of-the-art machine learning models such as ResNets \citep{he2016deep} or Transformers \citep{vaswani2017attention}.
Unfortunately, directly utilizing these large scale models is computationally infeasible, therefore we outer-train on proxy tasks for speed\citep{zoph2017learning}.

In order to outer-train a learned optimizer capable of generalizing to new optimization tasks, we utilize an outer-training task set consisting of around 6,000 tasks designed after~\citet{metz2020using}.
These tasks include RNNs~\citep{hochreiter1997long, chung2014empirical}, CNNs~\citep{lecun1998mnist}, masked auto regressive flows~\citep{papamakarios2017masked}, fully connected networks, language modeling, variational autoencoders~\citep{kingma2013auto}, simple 2D test functions, quadratic bowls, and more.
For tasks that require them, we additionally sample a dataset, batch size, network architecture, and initialization scheme.
To keep outer-training efficient, we ensure that all tasks take less than 100 milliseconds per-training step.
For each task that makes use of a dataset, we create four splits of the data to prevent leakage: training data, which we compute gradients on and use to update the inner-parameters;
inner-validation data, which is used to compute validation losses used by the learned optimizer;
outer-validation data, which is used to update the weights of the learned optimizer; 
and test data, which is used to test an already trained learned optimizer.
Because loss values vary in magnitude, when outer-training we normalize these outer-loss values by the best loss achieved by a baseline optimizer and the initial loss value. Note this normalization is not used during inner-training.

\subsection{Optimizer architecture}
\label{sec:outer_param}
\begin{figure}
    \centering
\begin{overpic}[width=\textwidth]{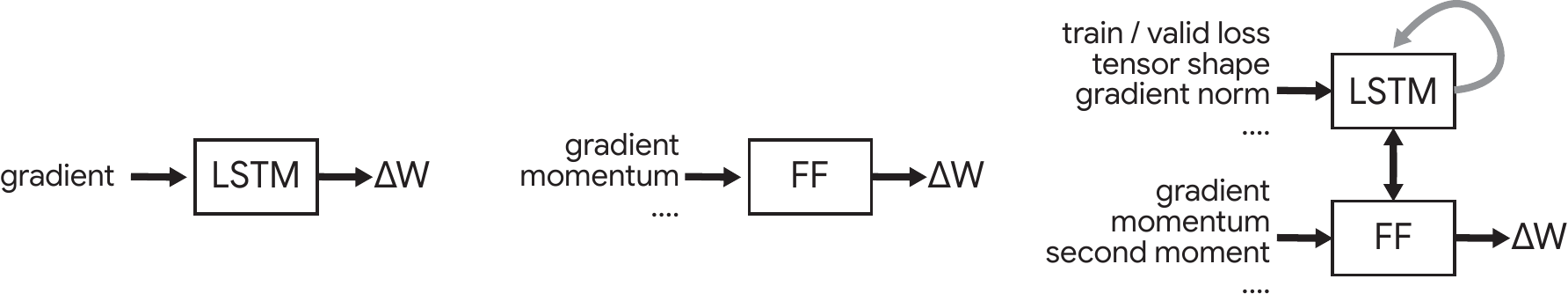}
 \put (0,18.0) {\textbf{\small(a)}}
 \put (32,18.0) {\textbf{\small(b)}}
 \put (63,18.0) {\textbf{\small(c)}}
\end{overpic}
    \caption{
    \textbf{(a)} The learned optimizer architecture proposed in \citet{andrychowicz2016learning} consisting of a per-parameter LSTM.
    \textbf{(b)} The learned optimizer architecture proposed in \citet{metz2019understanding} consisting of a per-parameter fully connected (feed-forward, FF) neural network with additional input features.
    \textbf{(c)} The learned optimizer architecture proposed in this work consisting of a per-tensor LSTM which exchanges information with a per-parameter feed forward neural network (FF). 
    The LSTMs associated with each tensor additionally share information with each other (shown in gray).
    \label{fig:archdiagram}
}
\end{figure}
Designing a learned optimizer architecture requires balancing computational efficiency and expressivity. Past work in learned optimizers has shown that incorporating inductive biases based on existing optimization techniques such as momentum or second moment accumulation leads to better performance \citep{wichrowska2017learned, metz2019understanding}. 
The optimizer we use in this work consists of a hierarchical optimizer similar to~\citep{wichrowska2017learned} (Figure \ref{fig:archdiagram}). A per-tensor LSTM is run on features computed over each parameter tensor. This LSTM then forwards information to the other tensors' LSTMs as well as to a per-parameter feedforward neural network. The per-parameter feedforward network additionally takes in information about gradients and parameter value, and outputs parameter updates. Additional outputs are aggregated and fed back into the per-tensor network. This information routing allows for communication across all components.

For per-parameter features we leverage effective inductive biases from hand-designed optimizers, and use a variety of features including the gradient, the parameter value, and momentum-like running averages of both. All features are normalized, in a fashion similar to that in RMSProp \citep{tieleman2012lecture}, or relative to the norm across the full tensor. 
For per-tensor features we use a variety of features built from summary statistics computed from the current tensor, the tensor's gradient, and running average features such as momentum and second moments. We also include information about the tensor's rank and shape.
We also feed global features into the per-tensor LSTM, such as training loss and validation loss, normalized so as to have a relatively consistent scale across tasks. 
To compute a weight update, the per-parameter MLP outputs two values, $(a,b)$, which are used to update inner-parameters: $w^{t+1}=w^{t}+\exp(a)b$.
See Appendix \ref{app:learned_opt_details} for many more details. 

\section{Results}

\subsection{Comparing learned optimizer architectures and training task set sizes}
\begin{figure}
    \centering
\begin{overpic}[width=0.45\textwidth]{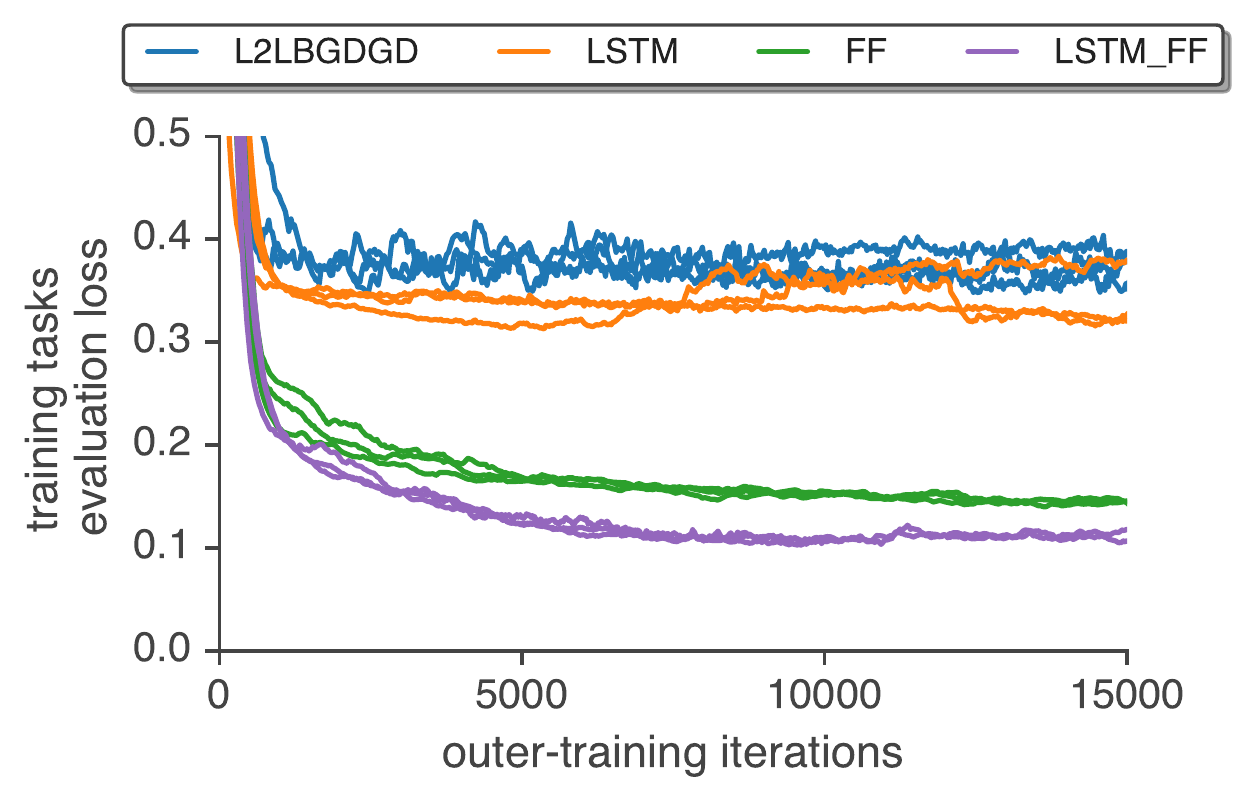}
 \put (0,58) {\textbf{\small(a)}}
\end{overpic}
\qquad
\begin{overpic}[width=0.45\textwidth]{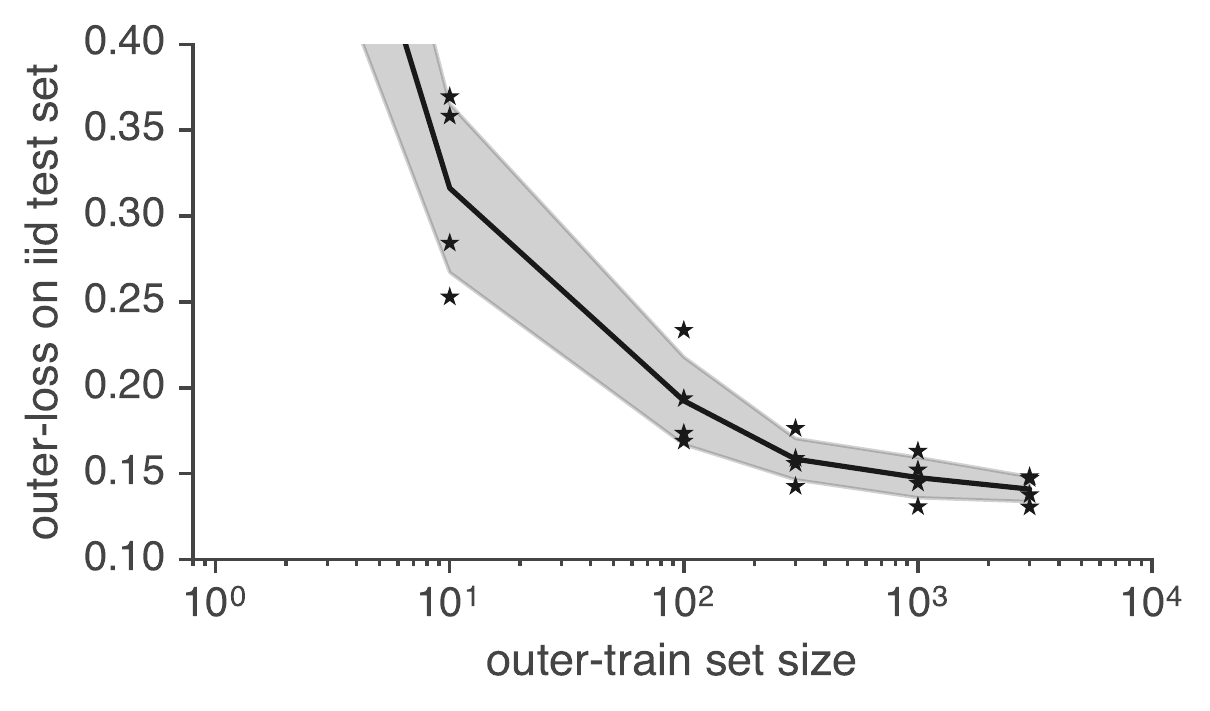}
 \put (0,58) {\textbf{\small(b)}}
\end{overpic}    
    \caption{
    {\textbf{(a)}} Our proposed learned optimizer has a greater sample efficiency than existing methods.
    On the x-axis we show outer-updates performed to outer-train each learned optimizer. 
    On the y-axis we show outer-loss. Each point consists of using a single outer parameter value to train 100 models for 10k inner-iterations, each with five random initializations. We show the average validation performance post-normalization averaged over all tasks, seeds, and inner-training steps.
    {\textbf{(b)}} Outer-generalization performance increases with an increasing number of outer-training tasks. We show the mean performance achieved after $\sim$10000 outer-training iterations (between 10500 and 9500) averaged over four random seeds while varying outer training set size. We show each seed as a star, and standard deviation as the shaded region. We find poor outer-generalization for smaller numbers of tasks. As we increase the number of outer-training tasks performance becomes more consistent, and lower loss values are achieved.
    \label{fig:meta_train}
    }
\end{figure}
First, we show experiments comparing the performance of different learned optimizer architectures from the literature.
We trained: a component-wise LSTM optimizer from \citet{andrychowicz2016learning} (L2LBGDGD), a modification of this LSTM with the decomposed direction and magnitude output from ~\citet{metz2019understanding} (LSTM), the fully connected optimizer from \citet{metz2019understanding} (FF), as well as the proposed learned optimizer in this work (\sref{sec:outer_param}) (LSTM\_FF).
As shown in Figure \ref{fig:meta_train}(a), the proposed architecture achieves the lowest outer-training loss and achieves this in the fewest outer-training steps.
To the best of our knowledge, this is the first published comparison across \textit{different} learned optimizer architectures, on the same suite of tasks. Previous work only compared a proposed learned optimizer against hand-designed (baseline) optimizers.

Next, we explored how increasing the number of inner-tasks used when training an optimizer affects final performance.
To do this, we randomly sampled subsets of tasks from the full task set, while evaluating performance on a common held-out set of tasks.
Figure~\ref{fig:meta_train}(b) shows that increasing the number of tasks leads to large improvements.

\subsection{Comparisons with hand-designed optimizers}

\begin{figure}
    \centering
    \includegraphics[width=\textwidth]{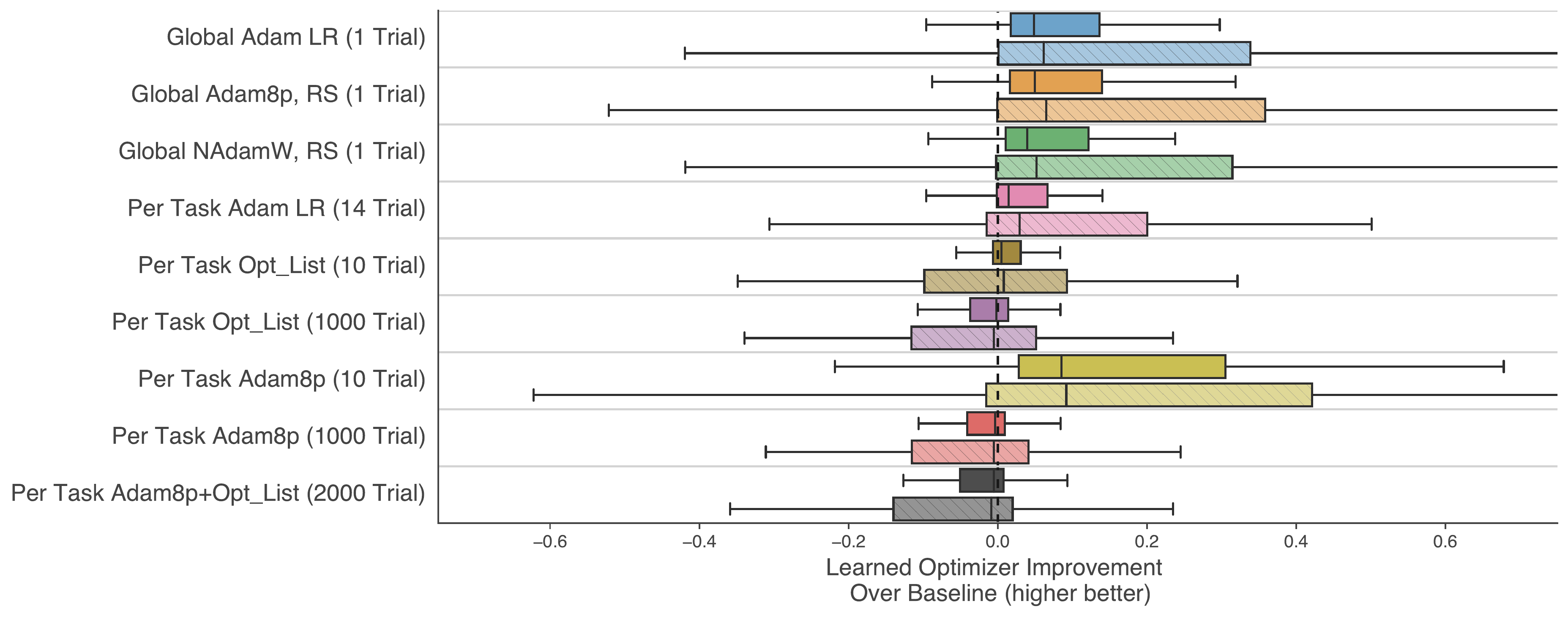}

    \caption{Learned optimizer performance as compared to a battery of different baselines. For each baseline, we show a box plot over 100 different tasks. We show both a training set (solid color) which the optimizer has seen, and a test set (hatched light color) which is sampled from the same distribution but never used for outer-training.
    Values to the right of the dashed line indicate the learned optimizer outperforming the corresponding baseline. The learned optimizer outperforms any single baseline optimizer with a fixed hyperparameter configuration (marked global, with a single Trial). 
    We also outperform per-task tuning of baselines, when tuning is done only over a modest number of hyperparameter configurations (for instance, learning rate tuned Adam which uses 14 trials). 
    Note that the learned optimizer undergoes no per-task tuning and only makes use of one trial.
    \label{fig:perf}
    }
\end{figure}

We compare against three baseline optimizers: \textbf{AdamLR}, which is the Adam optimizer~\cite{kingma2014adam} with a tuned learning rate. \textbf{Adam8p}, which is a version of the Adam optimizer with eight tunable hyperparameters: learning rate, $\beta_1$, $\beta_2$, and $\epsilon$, plus additional $\ell_1$ and $\ell_2$ regularization penalties, and a learning rate schedule parameterized with a linear decay and exponential decay.  See Appendix \ref{app:adam8p} for more details. Our final baseline optimizer, called \textbf{opt\_list}, consists of the NAdam optimizer~\citep{dozat2016incorporating} with ``AdamW'' style weight decay~\citep{loshchilov2017decoupled}, cosine learning rate schedules~\citep{loshchilov2016sgdr} and learning rate warm up (See \citet{metz2020using} for more info). Instead of tuning these with some search procedure, however, we draw them from a sorted list of hyperparameters provided by \citep{metz2020using} for increased sample efficiency.

Evaluation of optimizers, let alone learned optimizers, is difficult due to different tasks of interest, hyperparameter search strategies, and compute budgets~\citep{choi2019empirical, sivaprasad2019tunability}.
We structure our comparison by exploring two scenarios for how a machine learning practitioner might go about tuning the hyperparameters of an optimizer. 
First, we consider an ``off-the-shelf'' limit, where a practitioner performs a small number of optimizer evaluations using off-the-shelf methods (for instance, tuning learning rate only). This is typically done during exploration of a new machine learning model or dataset.
Second, we consider a ``finely tuned'' limit, where a practitioner has a large compute budget devoted to tuning hyperparameters of a traditional optimizer for a particular problem of interest.

For the ``off-the-shelf'' limit, we consider the scenario where a practitioner has access to a limited number of optimization runs (trials) ($\leq10$) for a particular problem. 
Thus, we select a first set of baseline hyperparameters using a single, default value (denoted \textit{global} in Fig~\ref{fig:perf}) across all of the tasks. We use random search (RS) using 1000 different hyperparameter values to find the global value that performs best on average for all tasks.

Practitioners often tune models with a small number of evaluations. As such, we include comparisons to per-task tuned learning rate tuned Adam, the first 10 entries of opt\_list, and 10 hyperparameter evaluations obtained from random search using the adam8p hyperparmeterization.

For the ``finely tuned'' limit, we consider task-specific hyperparameter tuning, where the hyperparameters for each baseline optimizer are selected individually for each task (denoted \textit{per-task} in Fig~\ref{fig:perf}).

We plot a histogram over tasks showing the difference in performance between the each baseline optimizer and the learned optimizer in each row of Fig~\ref{fig:perf}. First, we note that the distribution is broad, indicating that for some tasks the learned optimizer is much better, whereas for others, the baseline optimizer(s) are better. On average, we see small but consistent performance improvements over baseline optimizers, especially in the ``off-the-shelf'' scenario. We attribute this to the diverse set of tasks used for training the learned optimizer.

\subsection{Understanding optimizer behavior}

\begin{figure}
    \centering
    \includegraphics[width=4.5in]{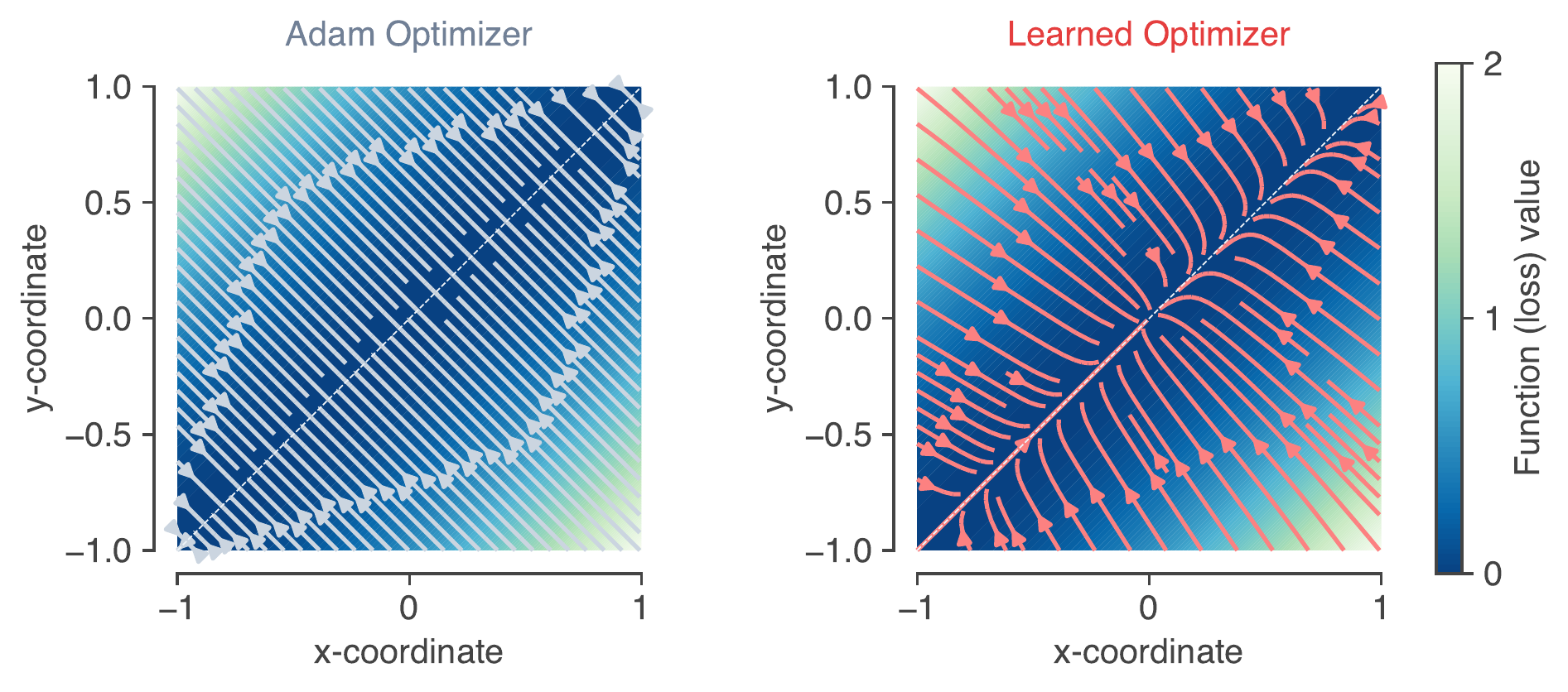}

    \caption{Implicit regularization in the learned optimizer. Panels show optimization trajectories over a 2D loss surface, $f(x, y) = \frac{1}{2}\left(x-y\right)^2$, that has a continuum of solutions along the diagonal. \textbf{Left:} Trajectories of the Adam optimizer go to the nearest point on the diagonal. \textbf{Right:} Trajectories of the learned optimizer go towards the diagonal, but also decay towards a single solution at the origin, \textit{as if} there were an added regularization penalty on the magnitude of each parameter.}
    \label{fig:implicit_reg}
\end{figure}

To better understand the behavior of the learned optimizer, we performed probe experiments where we compared trajectories of the learned optimizer on simple loss surfaces against baseline optimizers. The goal of these experiments was to generate insight into \textit{what} the learned optimizer has learned.

In machine learning, many problems benefit from including some kind of regularization penalty, such as an $\ell_2$ penalty on the weights or parameters of a model. We explored whether the learned optimizer (which was trained to minimize validation loss) had any \textit{implicit} regularization, beyond what was specified as part of the loss. To test this, we ran optimizers on a simple 2D loss surface, with a continuum of solutions along the diagonal: $f(x, y) = \frac{1}{2}\left(x-y\right)^2$. Although any point along the $x=y$ diagonal is a global minimum, we wanted to see if the learned optimizer would prefer any particular solution within that set.

Figure~\ref{fig:implicit_reg} shows the resulting training trajectories along the 2D loss surface from many starting points. For a baseline optimizer (left), the trajectories find the nearest point on the diagonal. However, we find that the learned optimizer has learned some implicit regularization, in that it pushes the parameters towards a solution with small norm: $(0,0)$.

\subsection{Generalization along different task axes}

\begin{figure}
    \centering
\begin{overpic}[width=\textwidth]{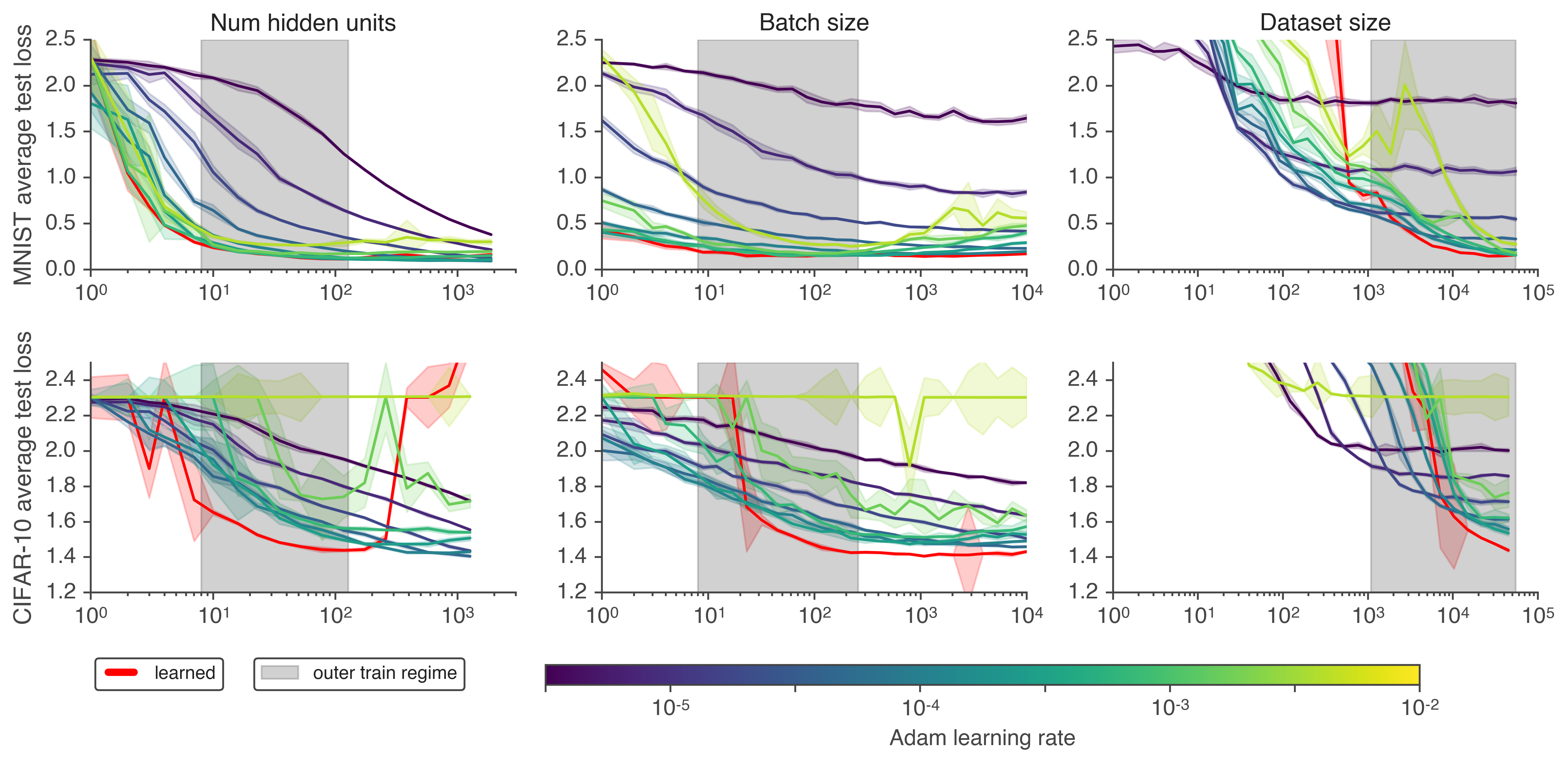}
\end{overpic}
    \caption{ 
    We show outer-generalization of the learned optimizer in a controlled setting by varying hyperparameters of the task being trained. We vary parameters around two types of models. \textbf{Top row:} A two hidden layer, 32 unit fully connected network trained on MNIST with batch size 128. \textbf{Bottom row:} A two layer hidden layer, 64 unit fully connected network trained on CIFAR-10 with batch size 128. We vary the width of the underlying network, batch size, and dataset size in each column.
    Each point on each curve shows the mean test loss averaged over 10k inner steps.
    We show median performance over five random task initializations. Error bars denote one standard deviation away from the median.
    Color denotes different learning rates for Adam log spaced every half order of magnitude for Adam with purple representing $10^{-6}$, and yellow at $10^{-2}$.
    We find the learned optimizer is able to generalize outside of the outer-training distribution (indicated by the shaded patch) in some cases.
    \label{fig:generalization_sweep}
    }
\end{figure}

Next, we wondered whether the learned optimizer was capable of training machine learning models which differed across different architectural and training hyperparameters.
To test this type of generalization, we trained fully connected neural networks on CIFAR-10 and MNIST, and swept three model hyperparameters: the number of hidden units per layer (network width), the batch size, and the number of training examples (dataset size, formed by subsampling the full dataset).
For each sweep, we compare the learned optimizer to a baseline optimizer, Adam, over a grid of eight different learning rates logarithmically spaced from $10^{-5.5}$ to $10^{-2}$.

The results of these experiments are in Fig.~\ref{fig:generalization_sweep}. 
As we vary the number of hidden units (left column) or batch size (middle column), the learned optimizer generalizes outside of the range of hidden units used during training (indicated by the shaded regions).
In addition, the learned optimizer matches the performance of the best learning rate tuned Adam optimizer.
On CIFAR-10, as we move further away from the outer-training task distribution, the learned optimizer diverges. 
For dataset size (right column), we find that the learned optimizer is more sensitive to the amount of data present (performance drops off more quickly as the dataset size decreases).
These experiments demonstrate the learned optimizer is capable of adapting to some aspects of target tasks which differ from its outer-training distribution, without additional tuning.

\subsection{Generalization to large-scale problems}

\begin{figure}
    \vspace{0.2em}
    \centering
\begin{overpic}[width=0.32\textwidth]{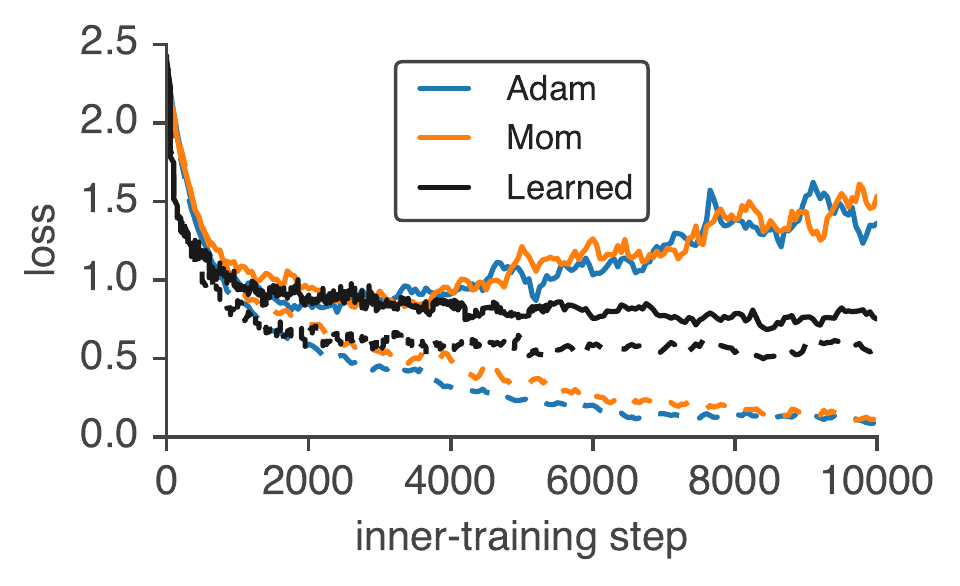}
 \put (0,62) {\textbf{\small(a)}}
\end{overpic}    
\begin{overpic}[width=0.32\textwidth]{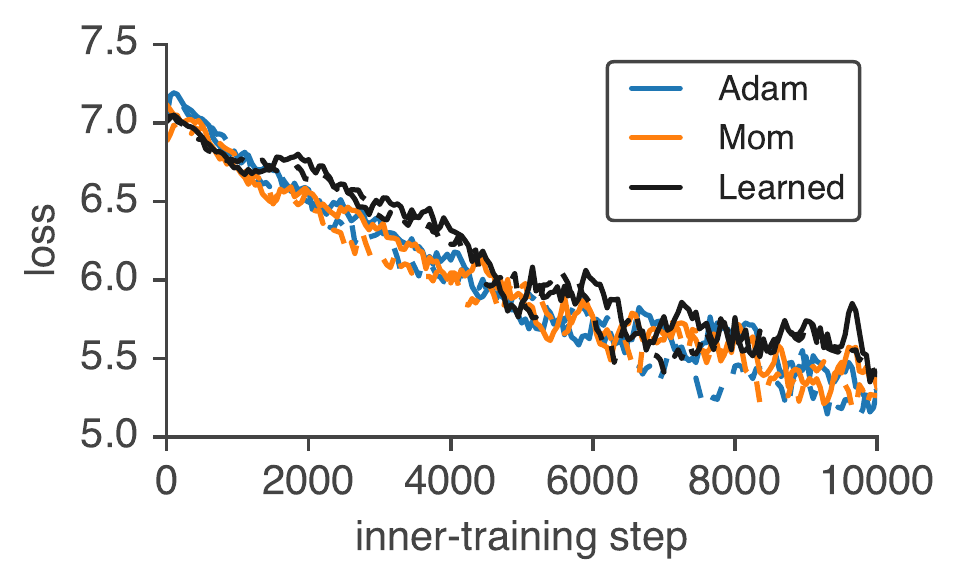}
 \put (0,62) {\textbf{\small(b)}}
\end{overpic}    
\begin{overpic}[width=0.32\textwidth]{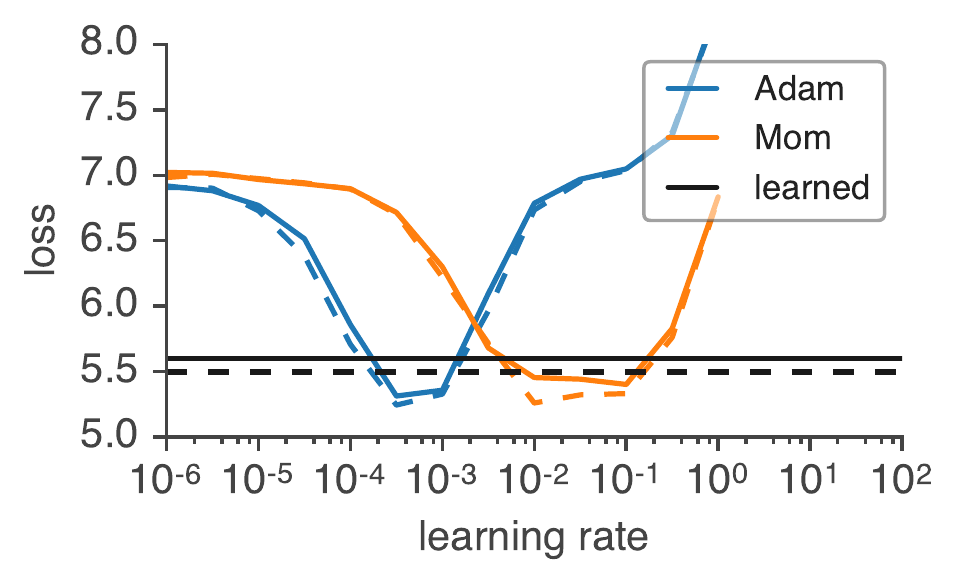}
 \put (0,62) {\textbf{\small(c)}}
\end{overpic}
    \caption{Learned optimizers are able to optimize ResNet models. 
    \textbf{(a)} Inner-learning curves using the learned optimizer to train a small ResNet on CIFAR-10. We compare to learning rate tuned Adam and Momentum. Solid lines denote test performance, dashed lines are train. 
    \textbf{(b)} Inner-learning curves using the learned optimizer to train a small ResNet on 64x64 resized ImageNet. We compare to learning rate tuned Adam and Momentum. 
    \textbf{(c)} Performance averaged between 9.5k and 10k inner-iterations for Adam and Momentum as a function of learning rate, on the same task as in (b). Despite requiring no tuning, the learned optimizer performs similarly to these baselines after tuning.
    \label{fig:large_resnet}
    }
\end{figure}

We test the learned optimizer on large-scale machine learning tasks.
We use two ResNet V2~\citep{he2016identity} architectures: a 14 layer residual network trained on CIFAR-10~\citep{krizhevsky2009cifar}; and a 35 layer residual network trained on 64x64 resized ImageNet~\citep{ILSVRC15}.
We train both networks with the learned optimizer and compare the performance to learning rate tuned Adam and Momentum~(Fig.~\ref{fig:large_resnet}). Details in Appendix~\ref{app:adam8p}.
For CIFAR-10, we find that the learned optimizer achieves similar performance as the baselines but does not overfit later in inner-training.
For ImageNet, we find that the learned optimizer performs slightly worse.

Note that our baselines only include learning rate tuning.
More specialized hyperparameter configurations, designed specifically for these tasks, such as learning rate schedules and data augmentation strategies will perform better.
An extensive study of learned optimizer performance on a wide range of state-of-the-art models is a subject for future work.

\subsection{Learned optimizers training themselves}

\begin{figure}
    \centering
    \includegraphics[width=0.45\textwidth]{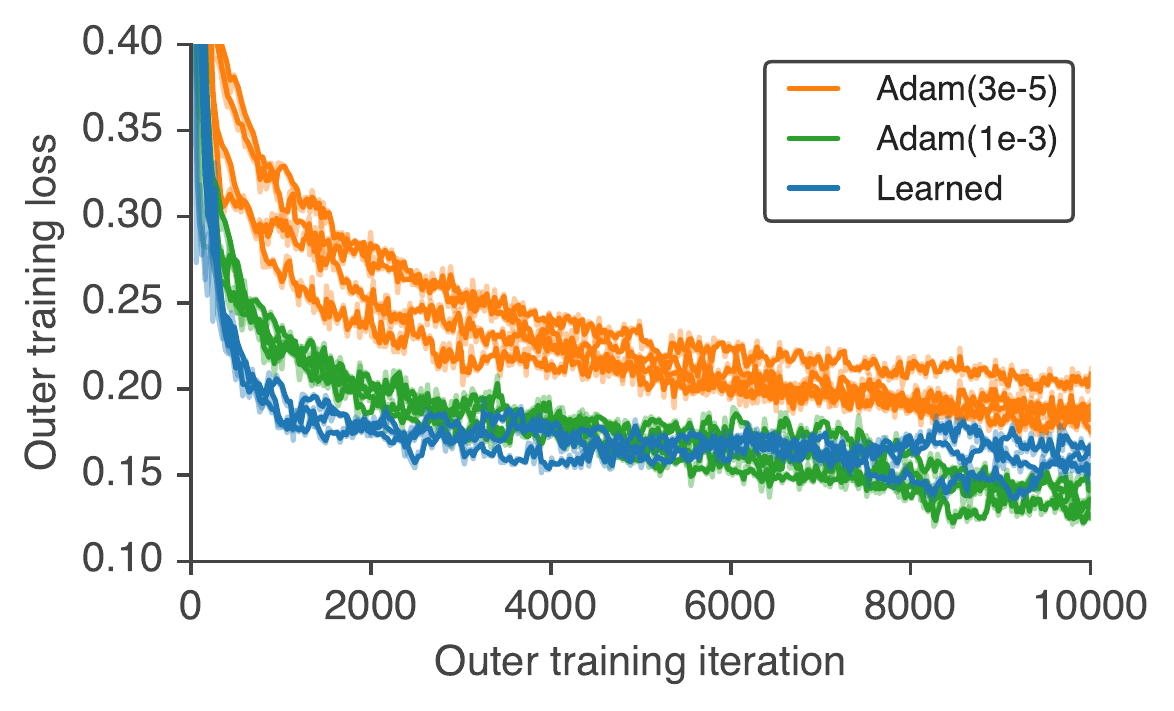}
    \caption{The learned optimizers can be used to train themselves about as efficiently as hand designed methods. On the x-axis we show number of weight updates done to the learned optimizer. On the y-axis we show outer-loss. Each point consists of inner-training 100 models, each with five random initializations trained for 10k inner-iterations. We show the average validation performance post-normalization averaged over all tasks, seeds, and inner-training steps. Each line represents a different randomly initialized learned optimizer. In orange we show Adam with a learning rate of 3e-5 the same value used to train the optimizers in this work. In green, we show Adam with a learning rate of 1e-3, the learning rate that performed best in this 10k outer-iteration regime. In blue we show our learned optimizer.
    }
    \label{fig:self_opt}
\end{figure}

Finally, we performed an experiment to test if a learned optimizer can be use to train new learned optimizers.
Figure \ref{fig:self_opt} shows that this ``self-optimized'' training curve is similar to the training curve using our hand-tuned training setup (using the Adam optimizer).
We interpret this as evidence of unexpectedly effective generalization, as the training of a learned optimizer is unlike anything in the set of training tasks used to train the optimizer.
We show outer-training for 10k outer-training iterations, matching the number of inner-iterations used when outer-training the learned optimizer. 
If outer-training is continued beyond 10k iterations, learned optimizer performance worsens (see Appendix~\ref{app:selfopt}), suggesting that more inner-iterations are needed when outer-training.

\section{Discussion}
In this work, we train a learned optimizer using a larger and more diverse set of training tasks, better optimization techniques, an improved optimizer architecture, and more compute than previous work. 
The resulting optimizer outperforms hand-designed optimizers constrained to a single set of hyperparameters, and performs comparably to hand designed optimizers after a modest hyperparameter search. Further, it demonstrates some ability to generalize to optimization tasks unlike those it was trained on. Most dramatically, it demonstrates an ability to train itself.

However, the learned optimizer we develop is not without limitations. Below we summarize areas for future work.

\textbf{Generalization:} While it performs well for tasks like those in TaskSet, we do not yet fully understand its outer-generalization capabilities. 
It shows promising ability to generalize to out of distribution problems such as training itself, and ResNet models, but it does not outperform competing algorithms in all settings on simple ``out of distribution'' optimization tasks, like those seen in Figure~\ref{fig:generalization_sweep}.

\textbf{Optimization / Compute:} Currently it takes significant compute expenditure to train a learned optimizer at this scale, resulting in a nontrivial carbon footprint. Ideally training can be run once and the resulting learned optimizer can be used broadly as is the case with BERT~\citep{devlin2018bert}.

\textbf{Learned optimizer architectures:} We have shown there is considerable improvement to be had by incorporating better inductive biases both in outer-optimization speed and capacity. Improving these architectures, and leveraging additional features will hopefully lead to more performant learned optimizers. Additionally, the current optimizer, while compute efficient, is memory inefficient requiring $>\!5\times$ more storage per-parameter than Adam. We do not believe this poses a fundamental barrier, but modifications similar to those in~\citet{shazeer2018adafactor} will be needed to train larger models.

\textbf{Outer-training task distribution:} We have shown that training on larger tasksets leads to better generalization. We have not studied which tasks should be included so as to aid outer-generalization.

\section*{Acknowledgements}
We would like to thank Alex Alemi, Eric Jang, Diogo Moitinho de Almeida, Timothy Nguyen, Alec Radford, Ruoxi Sun, Paul Vicol and Wojciech Zaremba for discussion related to this work as well as the Brain Team for providing a supportive research environment. We would also like to thank the authors of Numpy~\citep{oliphant2006guide, van2011numpy, harris2020array}, Seaborn~\citep{waskom2020seaborn}, and Matplotlib~\citep{matplotlib}. 

\section*{Broader Impact}
Machine learning training and inference is a major energy consumer, and training models likely dominates most individual machine learning researcher's carbon emissions~\citep{strubell2019energy}. 
By meta-learning optimizers, we hope to amortize the cost of training ML models, and thus reduce the environmental impact of training a single model. 
We hope to achieve this both by reducing the need for extensive hyperparameter tuning as models are developed, and making training more efficient for the best hyperparameters.

With this work, and more generally with meta-learning algorithms, we hope to provide researchers access to more performant easier to use optimizers for their problems. Improving technology to do machine learning will accelerate its impact, for better or worse. We believe machine learning technologies will be beneficial to humanity on the whole, and thus by improving the ability to optimize models we are moving towards this goal.

Our optimizer is trained on a mixture of different tasks and datasets. It is our goal to construct an outer-training distribution that learns relevant inductive biases for machine learning tasks of interest. However, we currently do not have good methods to decipher exactly what types of inductive biases a learned optimizer might learn. Thus, future work is needed to explore what kinds of information is absorbed by learned optimizers. In the meantime, care should be taken when employing learned optimizers as the inductive biases of a particular learned optimizer may not be appropriate for an end user's goals.

In fact, it has been suggested that learned optimizers could be a source of alignment drift in the development of Artificial General Intelligence (AGI) -- leading to AGI that does not perform as its creators hoped~\citep{hubinger2019risks}. While we do not feel that is an immediate danger from the present work, it is a consideration that should be kept in mind as learned optimizers become increasingly powerful.

\clearpage

\bibliography{references}

\begin{thebibliography}{77}
\providecommand{\natexlab}[1]{#1}
\providecommand{\url}[1]{\texttt{#1}}
\expandafter\ifx\csname urlstyle\endcsname\relax
  \providecommand{\doi}[1]{doi: #1}\else
  \providecommand{\doi}{doi: \begingroup \urlstyle{rm}\Url}\fi

\bibitem[Krizhevsky et~al.(2012)Krizhevsky, Sutskever, and
  Hinton]{krizhevsky2012imagenet}
Alex Krizhevsky, Ilya Sutskever, and Geoffrey~E Hinton.
\newblock Imagenet classification with deep convolutional neural networks.
\newblock In \emph{Advances in neural information processing systems}, pages
  1097--1105, 2012.

\bibitem[Berner et~al.(2019)Berner, Brockman, Chan, Cheung, D{\k{e}}biak,
  Dennison, Farhi, Fischer, Hashme, Hesse, et~al.]{berner2019dota}
Christopher Berner, Greg Brockman, Brooke Chan, Vicki Cheung, Przemys{\l}aw
  D{\k{e}}biak, Christy Dennison, David Farhi, Quirin Fischer, Shariq Hashme,
  Chris Hesse, et~al.
\newblock Dota 2 with large scale deep reinforcement learning.
\newblock \emph{arXiv preprint arXiv:1912.06680}, 2019.

\bibitem[Vinyals et~al.(2019)Vinyals, Babuschkin, Chung, Mathieu, Jaderberg,
  Czarnecki, Dudzik, Huang, Georgiev, Powell, et~al.]{vinyals2019alphastar}
Oriol Vinyals, Igor Babuschkin, Junyoung Chung, Michael Mathieu, Max Jaderberg,
  Wojciech~M Czarnecki, Andrew Dudzik, Aja Huang, Petko Georgiev, Richard
  Powell, et~al.
\newblock Alphastar: Mastering the real-time strategy game starcraft ii.
\newblock \emph{DeepMind blog}, page~2, 2019.

\bibitem[Piech et~al.(2015)Piech, Bassen, Huang, Ganguli, Sahami, Guibas, and
  Sohl-Dickstein]{piech2015deep}
Chris Piech, Jonathan Bassen, Jonathan Huang, Surya Ganguli, Mehran Sahami,
  Leonidas~J Guibas, and Jascha Sohl-Dickstein.
\newblock Deep knowledge tracing.
\newblock In \emph{Advances in neural information processing systems}, pages
  505--513, 2015.

\bibitem[Choi et~al.(2019)Choi, Shallue, Nado, Lee, Maddison, and
  Dahl]{choi2019empirical}
Dami Choi, Christopher~J Shallue, Zachary Nado, Jaehoon Lee, Chris~J Maddison,
  and George~E Dahl.
\newblock On empirical comparisons of optimizers for deep learning.
\newblock \emph{arXiv preprint arXiv:1910.05446}, 2019.

\bibitem[Daniel et~al.(2016)Daniel, Taylor, and Nowozin]{daniel2016learning}
Christian Daniel, Jonathan Taylor, and Sebastian Nowozin.
\newblock Learning step size controllers for robust neural network training.
\newblock In \emph{Thirtieth AAAI Conference on Artificial Intelligence}, 2016.

\bibitem[Xu et~al.(2017)Xu, Qin, Wang, and Liu]{xu2017reinforcement}
Chang Xu, Tao Qin, Gang Wang, and Tie-Yan Liu.
\newblock Reinforcement learning for learning rate control.
\newblock \emph{arXiv preprint arXiv:1705.11159}, 2017.

\bibitem[Xu et~al.(2019)Xu, Dai, Kemp, and Metz]{xu2019learning}
Zhen Xu, Andrew~M Dai, Jonas Kemp, and Luke Metz.
\newblock Learning an adaptive learning rate schedule.
\newblock \emph{arXiv preprint arXiv:1909.09712}, 2019.

\bibitem[Andrychowicz et~al.(2016)Andrychowicz, Denil, Gomez, Hoffman, Pfau,
  Schaul, and de~Freitas]{andrychowicz2016learning}
Marcin Andrychowicz, Misha Denil, Sergio Gomez, Matthew~W Hoffman, David Pfau,
  Tom Schaul, and Nando de~Freitas.
\newblock Learning to learn by gradient descent by gradient descent.
\newblock In \emph{Advances in Neural Information Processing Systems}, pages
  3981--3989, 2016.

\bibitem[Wichrowska et~al.(2017)Wichrowska, Maheswaranathan, Hoffman,
  Colmenarejo, Denil, de~Freitas, and Sohl-Dickstein]{wichrowska2017learned}
Olga Wichrowska, Niru Maheswaranathan, Matthew~W Hoffman, Sergio~Gomez
  Colmenarejo, Misha Denil, Nando de~Freitas, and Jascha Sohl-Dickstein.
\newblock Learned optimizers that scale and generalize.
\newblock \emph{International Conference on Machine Learning}, 2017.

\bibitem[Lv et~al.(2017)Lv, Jiang, and Li]{lv2017learning}
Kaifeng Lv, Shunhua Jiang, and Jian Li.
\newblock Learning gradient descent: Better generalization and longer horizons.
\newblock \emph{arXiv preprint arXiv:1703.03633}, 2017.

\bibitem[Metz et~al.(2018)Metz, Maheswaranathan, Cheung, and
  Sohl-Dickstein]{metz2018learning}
Luke Metz, Niru Maheswaranathan, Brian Cheung, and Jascha Sohl-Dickstein.
\newblock Learning unsupervised learning rules.
\newblock \emph{arXiv preprint arXiv:1804.00222}, 2018.

\bibitem[Metz et~al.(2019{\natexlab{a}})Metz, Maheswaranathan, Shlens,
  Sohl-Dickstein, and Cubuk]{metz2019using}
Luke Metz, Niru Maheswaranathan, Jonathon Shlens, Jascha Sohl-Dickstein, and
  Ekin~D Cubuk.
\newblock Using learned optimizers to make models robust to input noise.
\newblock \emph{arXiv preprint arXiv:1906.03367}, 2019{\natexlab{a}}.

\bibitem[Metz et~al.(2019{\natexlab{b}})Metz, Maheswaranathan, Nixon, Freeman,
  and Sohl-Dickstein]{metz2019understanding}
Luke Metz, Niru Maheswaranathan, Jeremy Nixon, Daniel Freeman, and Jascha
  Sohl-Dickstein.
\newblock Understanding and correcting pathologies in the training of learned
  optimizers.
\newblock In \emph{International Conference on Machine Learning}, pages
  4556--4565, 2019{\natexlab{b}}.

\bibitem[Gu et~al.(2019)Gu, Greydanus, Metz, Maheswaranathan, and
  Sohl-Dickstein]{gu2019meta}
Keren Gu, Sam Greydanus, Luke Metz, Niru Maheswaranathan, and Jascha
  Sohl-Dickstein.
\newblock Meta-learning biologically plausible semi-supervised update rules.
\newblock \emph{bioRxiv}, 2019.

\bibitem[Metz et~al.(2020)Metz, Maheswaranathan, Sun, Freeman, Poole, and
  Sohl-Dickstein]{metz2020using}
Luke Metz, Niru Maheswaranathan, Ruoxi Sun, C~Daniel Freeman, Ben Poole, and
  Jascha Sohl-Dickstein.
\newblock Using a thousand optimization tasks to learn hyperparameter search
  strategies.
\newblock \emph{arXiv preprint arXiv:2002.11887}, 2020.

\bibitem[Hart and Levin(1962)]{hart1962ai}
M~Levin~T Hart and Mike Levin.
\newblock Ai memo 39-the new compiler.
\newblock Technical report, Technical report, MIT, 1962.

\bibitem[Franceschi et~al.(2018)Franceschi, Frasconi, Salzo, and
  Pontil]{franceschi2018bilevel}
Luca Franceschi, Paolo Frasconi, Saverio Salzo, and Massimilano Pontil.
\newblock Bilevel programming for hyperparameter optimization and
  meta-learning.
\newblock \emph{arXiv preprint arXiv:1806.04910}, 2018.

\bibitem[Wu et~al.(2016)Wu, Ren, Liao, and Grosse]{wuunderstanding}
Yuhuai Wu, Mengye Ren, Renjie Liao, and Roger~B Grosse.
\newblock Understanding short-horizon bias in stochastic meta-optimization.
\newblock pages 478--487, 2016.

\bibitem[Rechenberg(1973)]{rechenberg1973evolutionsstrategie}
Ingo Rechenberg.
\newblock Evolutionsstrategie--optimierung technisher systeme nach prinzipien
  der biologischen evolution.
\newblock 1973.

\bibitem[Vicol and et. al.(2020)]{pes}
Paul Vicol and et. al.
\newblock Persisted evolutionary strategies.
\newblock \emph{In Preparation}, 2020.

\bibitem[He et~al.(2016{\natexlab{a}})He, Zhang, Ren, and Sun]{he2016deep}
Kaiming He, Xiangyu Zhang, Shaoqing Ren, and Jian Sun.
\newblock Deep residual learning for image recognition.
\newblock In \emph{Proceedings of the IEEE conference on computer vision and
  pattern recognition}, pages 770--778, 2016{\natexlab{a}}.

\bibitem[Vaswani et~al.(2017)Vaswani, Shazeer, Parmar, Uszkoreit, Jones, Gomez,
  Kaiser, and Polosukhin]{vaswani2017attention}
Ashish Vaswani, Noam Shazeer, Niki Parmar, Jakob Uszkoreit, Llion Jones,
  Aidan~N Gomez, {\L}ukasz Kaiser, and Illia Polosukhin.
\newblock Attention is all you need.
\newblock In \emph{Advances in neural information processing systems}, pages
  5998--6008, 2017.

\bibitem[Zoph et~al.(2018)Zoph, Vasudevan, Shlens, and Le]{zoph2017learning}
Barret Zoph, Vijay Vasudevan, Jonathon Shlens, and Quoc~V Le.
\newblock Learning transferable architectures for scalable image recognition.
\newblock \emph{Proceedings of the IEEE conference on computer vision and
  pattern recognition}, 2018.

\bibitem[Hochreiter and Schmidhuber(1997)]{hochreiter1997long}
Sepp Hochreiter and J{\"u}rgen Schmidhuber.
\newblock Long short-term memory.
\newblock \emph{Neural computation}, 9\penalty0 (8):\penalty0 1735--1780, 1997.

\bibitem[Chung et~al.(2014)Chung, Gulcehre, Cho, and
  Bengio]{chung2014empirical}
Junyoung Chung, Caglar Gulcehre, KyungHyun Cho, and Yoshua Bengio.
\newblock Empirical evaluation of gated recurrent neural networks on sequence
  modeling.
\newblock \emph{arXiv preprint arXiv:1412.3555}, 2014.

\bibitem[LeCun(1998)]{lecun1998mnist}
Yann LeCun.
\newblock The mnist database of handwritten digits.
\newblock \emph{http://yann. lecun. com/exdb/mnist/}, 1998.

\bibitem[Papamakarios et~al.(2017)Papamakarios, Pavlakou, and
  Murray]{papamakarios2017masked}
George Papamakarios, Theo Pavlakou, and Iain Murray.
\newblock Masked autoregressive flow for density estimation.
\newblock In \emph{Advances in Neural Information Processing Systems}, pages
  2338--2347, 2017.

\bibitem[Kingma and Welling(2013)]{kingma2013auto}
Diederik~P Kingma and Max Welling.
\newblock Auto-encoding variational bayes.
\newblock \emph{arXiv preprint arXiv:1312.6114}, 2013.

\bibitem[Tieleman and Hinton(2012)]{tieleman2012lecture}
Tijmen Tieleman and Geoffrey Hinton.
\newblock Lecture 6.5-rmsprop: Divide the gradient by a running average of its
  recent magnitude.
\newblock \emph{COURSERA: Neural networks for machine learning}, 4\penalty0
  (2):\penalty0 26--31, 2012.

\bibitem[Kingma and Ba(2014)]{kingma2014adam}
Diederik~P Kingma and Jimmy Ba.
\newblock Adam: A method for stochastic optimization.
\newblock \emph{arXiv preprint arXiv:1412.6980}, 2014.

\bibitem[Dozat(2016)]{dozat2016incorporating}
Timothy Dozat.
\newblock Incorporating nesterov momentum into adam.
\newblock 2016.

\bibitem[Loshchilov and Hutter(2017)]{loshchilov2017decoupled}
Ilya Loshchilov and Frank Hutter.
\newblock Decoupled weight decay regularization.
\newblock \emph{arXiv preprint arXiv:1711.05101}, 2017.

\bibitem[Loshchilov and Hutter(2016)]{loshchilov2016sgdr}
Ilya Loshchilov and Frank Hutter.
\newblock Sgdr: Stochastic gradient descent with warm restarts.
\newblock \emph{arXiv preprint arXiv:1608.03983}, 2016.

\bibitem[Sivaprasad et~al.(2019)Sivaprasad, Mai, Vogels, Jaggi, and
  Fleuret]{sivaprasad2019tunability}
Prabhu~Teja Sivaprasad, Florian Mai, Thijs Vogels, Martin Jaggi, and
  Fran{\c{c}}ois Fleuret.
\newblock On the tunability of optimizers in deep learning.
\newblock \emph{arXiv preprint arXiv:1910.11758}, 2019.

\bibitem[He et~al.(2016{\natexlab{b}})He, Zhang, Ren, and Sun]{he2016identity}
Kaiming He, Xiangyu Zhang, Shaoqing Ren, and Jian Sun.
\newblock Identity mappings in deep residual networks.
\newblock In \emph{European conference on computer vision}, pages 630--645.
  Springer, 2016{\natexlab{b}}.

\bibitem[Krizhevsky et~al.(2009)Krizhevsky, Nair, and
  Hinton]{krizhevsky2009cifar}
Alex Krizhevsky, Vinod Nair, and Geoffrey Hinton.
\newblock Cifar-10 and cifar-100 datasets.
\newblock \emph{URl: https://www. cs. toronto. edu/kriz/cifar. html}, 6, 2009.

\bibitem[Russakovsky et~al.(2015)Russakovsky, Deng, Su, Krause, Satheesh, Ma,
  Huang, Karpathy, Khosla, Bernstein, Berg, and Fei-Fei]{ILSVRC15}
Olga Russakovsky, Jia Deng, Hao Su, Jonathan Krause, Sanjeev Satheesh, Sean Ma,
  Zhiheng Huang, Andrej Karpathy, Aditya Khosla, Michael Bernstein,
  Alexander~C. Berg, and Li~Fei-Fei.
\newblock {ImageNet Large Scale Visual Recognition Challenge}.
\newblock \emph{International Journal of Computer Vision (IJCV)}, 115\penalty0
  (3):\penalty0 211--252, 2015.
\newblock \doi{10.1007/s11263-015-0816-y}.

\bibitem[Devlin et~al.(2018)Devlin, Chang, Lee, and Toutanova]{devlin2018bert}
Jacob Devlin, Ming-Wei Chang, Kenton Lee, and Kristina Toutanova.
\newblock Bert: Pre-training of deep bidirectional transformers for language
  understanding.
\newblock \emph{arXiv preprint arXiv:1810.04805}, 2018.

\bibitem[Shazeer and Stern(2018)]{shazeer2018adafactor}
Noam Shazeer and Mitchell Stern.
\newblock Adafactor: Adaptive learning rates with sublinear memory cost.
\newblock \emph{arXiv preprint arXiv:1804.04235}, 2018.

\bibitem[Oliphant(2006)]{oliphant2006guide}
Travis~E Oliphant.
\newblock \emph{A guide to NumPy}, volume~1.
\newblock Trelgol Publishing USA, 2006.

\bibitem[Van Der~Walt et~al.(2011)Van Der~Walt, Colbert, and
  Varoquaux]{van2011numpy}
Stefan Van Der~Walt, S~Chris Colbert, and Gael Varoquaux.
\newblock The numpy array: a structure for efficient numerical computation.
\newblock \emph{Computing in Science \& Engineering}, 13\penalty0 (2):\penalty0
  22, 2011.

\bibitem[Harris et~al.(2020)Harris, Millman, van~der Walt, Gommers, Virtanen,
  Cournapeau, Wieser, Taylor, Berg, Smith, et~al.]{harris2020array}
Charles~R Harris, K~Jarrod Millman, St{\'e}fan~J van~der Walt, Ralf Gommers,
  Pauli Virtanen, David Cournapeau, Eric Wieser, Julian Taylor, Sebastian Berg,
  Nathaniel~J Smith, et~al.
\newblock Array programming with numpy.
\newblock \emph{Nature}, 585\penalty0 (7825):\penalty0 357--362, 2020.

\bibitem[Waskom and the seaborn~development team(2020)]{waskom2020seaborn}
Michael Waskom and the seaborn~development team.
\newblock mwaskom/seaborn, September 2020.
\newblock URL \url{https://doi.org/10.5281/zenodo.592845}.

\bibitem[Hunter(2007)]{matplotlib}
J.~D. Hunter.
\newblock Matplotlib: A 2d graphics environment.
\newblock \emph{Computing in Science \& Engineering}, 9\penalty0 (3):\penalty0
  90--95, 2007.
\newblock \doi{10.1109/MCSE.2007.55}.

\bibitem[Strubell et~al.(2019)Strubell, Ganesh, and
  McCallum]{strubell2019energy}
Emma Strubell, Ananya Ganesh, and Andrew McCallum.
\newblock Energy and policy considerations for deep learning in nlp.
\newblock \emph{arXiv preprint arXiv:1906.02243}, 2019.

\bibitem[Hubinger et~al.(2019)Hubinger, van Merwijk, Mikulik, Skalse, and
  Garrabrant]{hubinger2019risks}
Evan Hubinger, Chris van Merwijk, Vladimir Mikulik, Joar Skalse, and Scott
  Garrabrant.
\newblock Risks from learned optimization in advanced machine learning systems.
\newblock \emph{arXiv preprint arXiv:1906.01820}, 2019.

\bibitem[Bello et~al.(2017)Bello, Zoph, Vasudevan, and Le]{Bello17}
Irwan Bello, Barret Zoph, Vijay Vasudevan, and Quoc Le.
\newblock Neural optimizer search with reinforcement learning.
\newblock 2017.
\newblock URL \url{https://arxiv.org/pdf/1709.07417.pdf}.

\bibitem[Bengio et~al.(1992)Bengio, Bengio, Cloutier, and
  Gecsei]{bengio1992optimization}
Samy Bengio, Yoshua Bengio, Jocelyn Cloutier, and Jan Gecsei.
\newblock On the optimization of a synaptic learning rule.
\newblock In \emph{Preprints Conf. Optimality in Artificial and Biological
  Neural Networks}, pages 6--8. Univ. of Texas, 1992.

\bibitem[Runarsson and Jonsson(2000)]{runarsson2000evolution}
Thomas~Philip Runarsson and Magnus~Thor Jonsson.
\newblock Evolution and design of distributed learning rules.
\newblock In \emph{Combinations of Evolutionary Computation and Neural
  Networks, 2000 IEEE Symposium on}, pages 59--63. IEEE, 2000.

\bibitem[Bengio et~al.(2013)Bengio, Boulanger-Lewandowski, and
  Pascanu]{bengio2013advances}
Yoshua Bengio, Nicolas Boulanger-Lewandowski, and Razvan Pascanu.
\newblock Advances in optimizing recurrent networks.
\newblock In \emph{2013 IEEE International Conference on Acoustics, Speech and
  Signal Processing}, pages 8624--8628. IEEE, 2013.

\bibitem[Graves et~al.(2014)Graves, Wayne, and Danihelka]{graves2014neural}
Alex Graves, Greg Wayne, and Ivo Danihelka.
\newblock Neural turing machines.
\newblock \emph{arXiv preprint arXiv:1410.5401}, 2014.

\bibitem[Nocedal(1980)]{nocedal1980updating}
Jorge Nocedal.
\newblock Updating quasi-newton matrices with limited storage.
\newblock \emph{Mathematics of computation}, 35\penalty0 (151):\penalty0
  773--782, 1980.

\bibitem[Liu and Nocedal(1989)]{liu1989limited}
Dong~C Liu and Jorge Nocedal.
\newblock On the limited memory bfgs method for large scale optimization.
\newblock \emph{Mathematical programming}, 45\penalty0 (1-3):\penalty0
  503--528, 1989.

\bibitem[Heess et~al.(2017)Heess, Sriram, Lemmon, Merel, Wayne, Tassa, Erez,
  Wang, Eslami, Riedmiller, et~al.]{heess2017emergence}
Nicolas Heess, Srinivasan Sriram, Jay Lemmon, Josh Merel, Greg Wayne, Yuval
  Tassa, Tom Erez, Ziyu Wang, SM~Eslami, Martin Riedmiller, et~al.
\newblock Emergence of locomotion behaviours in rich environments.
\newblock \emph{arXiv preprint arXiv:1707.02286}, 2017.

\bibitem[Cobbe et~al.(2018)Cobbe, Klimov, Hesse, Kim, and
  Schulman]{cobbe2018quantifying}
Karl Cobbe, Oleg Klimov, Chris Hesse, Taehoon Kim, and John Schulman.
\newblock Quantifying generalization in reinforcement learning.
\newblock \emph{arXiv preprint arXiv:1812.02341}, 2018.

\bibitem[Shahriari et~al.(2015)Shahriari, Swersky, Wang, Adams, and
  De~Freitas]{shahriari2015taking}
Bobak Shahriari, Kevin Swersky, Ziyu Wang, Ryan~P Adams, and Nando De~Freitas.
\newblock Taking the human out of the loop: A review of bayesian optimization.
\newblock \emph{Proceedings of the IEEE}, 104\penalty0 (1):\penalty0 148--175,
  2015.

\bibitem[Golovin et~al.(2017)Golovin, Solnik, Moitra, Kochanski, Karro, and
  Sculley]{golovin2017}
Daniel Golovin, Benjamin Solnik, Subhodeep Moitra, Greg Kochanski, John Karro,
  and D~Sculley.
\newblock Google vizier: A service for black-box optimization.
\newblock In \emph{International Conference on Knowledge Discovery and Data
  Mining}, 2017.

\bibitem[Li and Malik(2017{\natexlab{a}})]{li2016learning}
Ke~Li and Jitendra Malik.
\newblock Learning to optimize.
\newblock \emph{International Conference on Learning Representations},
  2017{\natexlab{a}}.

\bibitem[Li and Malik(2017{\natexlab{b}})]{li2017learning}
Ke~Li and Jitendra Malik.
\newblock Learning to optimize neural nets.
\newblock \emph{arXiv preprint arXiv:1703.00441}, 2017{\natexlab{b}}.

\bibitem[Schulman et~al.(2017)Schulman, Wolski, Dhariwal, Radford, and
  Klimov]{schulman2017proximal}
John Schulman, Filip Wolski, Prafulla Dhariwal, Alec Radford, and Oleg Klimov.
\newblock Proximal policy optimization algorithms.
\newblock \emph{arXiv preprint arXiv:1707.06347}, 2017.

\bibitem[Peters et~al.(2010)Peters, Mulling, and Altun]{peters2010relative}
Jan Peters, Katharina Mulling, and Yasemin Altun.
\newblock Relative entropy policy search.
\newblock In \emph{Twenty-Fourth AAAI Conference on Artificial Intelligence},
  2010.

\bibitem[Bengio(2000)]{bengio2000gradient}
Yoshua Bengio.
\newblock Gradient-based optimization of hyperparameters.
\newblock \emph{Neural computation}, 12\penalty0 (8):\penalty0 1889--1900,
  2000.

\bibitem[Baydin and Pearlmutter(2014)]{baydin2014automatic}
Atilim~Gunes Baydin and Barak~A Pearlmutter.
\newblock Automatic differentiation of algorithms for machine learning.
\newblock \emph{arXiv preprint arXiv:1404.7456}, 2014.

\bibitem[Maclaurin et~al.(2015)Maclaurin, Duvenaud, and
  Adams]{maclaurin2015gradient}
Dougal Maclaurin, David Duvenaud, and Ryan Adams.
\newblock Gradient-based hyperparameter optimization through reversible
  learning.
\newblock In \emph{International Conference on Machine Learning}, pages
  2113--2122, 2015.

\bibitem[Werbos(1990)]{werbos1990backpropagation}
Paul~J Werbos.
\newblock Backpropagation through time: what it does and how to do it.
\newblock \emph{Proceedings of the IEEE}, 78\penalty0 (10):\penalty0
  1550--1560, 1990.

\bibitem[Tallec and Ollivier(2017)]{tallec2017unbiasing}
Corentin Tallec and Yann Ollivier.
\newblock Unbiasing truncated backpropagation through time.
\newblock \emph{arXiv preprint arXiv:1705.08209}, 2017.

\bibitem[Finn et~al.(2017)Finn, Abbeel, and Levine]{finn2017model}
Chelsea Finn, Pieter Abbeel, and Sergey Levine.
\newblock Model-agnostic meta-learning for fast adaptation of deep networks.
\newblock \emph{arXiv preprint arXiv:1703.03400}, 2017.

\bibitem[Nichol et~al.(2018)Nichol, Achiam, and Schulman]{nichol2018first}
Alex Nichol, Joshua Achiam, and John Schulman.
\newblock On first-order meta-learning algorithms.
\newblock \emph{arXiv preprint arXiv:1803.02999}, 2018.

\bibitem[Pearlmutter(1996)]{pearlmutter1996investigation}
Barak Pearlmutter.
\newblock \emph{An investigation of the gradient descent process in neural
  networks}.
\newblock PhD thesis, Carnegie Mellon University Pittsburgh, PA, 1996.

\bibitem[Rastrigin(1963)]{rastrigin1963convergence}
LA~Rastrigin.
\newblock About convergence of random search method in extremal control of
  multi-parameter systems.
\newblock \emph{Avtomat. i Telemekh}, 24\penalty0 (11):\penalty0 1467--1473,
  1963.

\bibitem[Nesterov and Spokoiny(2011)]{nesterov2011random}
Yurii Nesterov and Vladimir Spokoiny.
\newblock Random gradient-free minimization of convex functions.
\newblock Technical report, Universit{\'e} catholique de Louvain, Center for
  Operations Research and Econometrics (CORE), 2011.

\bibitem[Choromanski et~al.(2018)Choromanski, Rowland, Sindhwani, Turner, and
  Weller]{choromanski2018structured}
Krzysztof Choromanski, Mark Rowland, Vikas Sindhwani, Richard~E Turner, and
  Adrian Weller.
\newblock Structured evolution with compact architectures for scalable policy
  optimization.
\newblock \emph{arXiv preprint arXiv:1804.02395}, 2018.

\bibitem[Salimans et~al.(2017)Salimans, Ho, Chen, Sidor, and
  Sutskever]{salimans2017evolution}
Tim Salimans, Jonathan Ho, Xi~Chen, Szymon Sidor, and Ilya Sutskever.
\newblock Evolution strategies as a scalable alternative to reinforcement
  learning.
\newblock \emph{arXiv preprint arXiv:1703.03864}, 2017.

\bibitem[Maheswaranathan et~al.(2019)Maheswaranathan, Metz, Tucker, Choi, and
  Sohl-Dickstein]{maheswaranathan2019guided}
Niru Maheswaranathan, Luke Metz, George Tucker, Dami Choi, and Jascha
  Sohl-Dickstein.
\newblock Guided evolutionary strategies: Augmenting random search with
  surrogate gradients.
\newblock In \emph{International Conference on Machine Learning}, pages
  4264--4273. PMLR, 2019.

\bibitem[Ghemawat et~al.(2003)Ghemawat, Gobioff, and Leung]{ghemawat2003google}
Sanjay Ghemawat, Howard Gobioff, and Shun-Tak Leung.
\newblock The google file system.
\newblock In \emph{Proceedings of the nineteenth ACM symposium on Operating
  systems principles}, pages 29--43, 2003.

\bibitem[Abadi et~al.(2016)Abadi, Barham, Chen, Chen, Davis, Dean, Devin,
  Ghemawat, Irving, Isard, et~al.]{abadi2016tensorflow}
Mart{\'\i}n Abadi, Paul Barham, Jianmin Chen, Zhifeng Chen, Andy Davis, Jeffrey
  Dean, Matthieu Devin, Sanjay Ghemawat, Geoffrey Irving, Michael Isard, et~al.
\newblock Tensorflow: A system for large-scale machine learning.
\newblock In \emph{OSDI}, volume~16, pages 265--283, 2016.

\end{thebibliography}
\bibliographystyle{unsrtnat}

\clearpage

\appendix

\section{Extended Related Work}
\label{app:mega_related_work}

We categorize progress in learned learned optimizers into three categories: parameterizations, data distributions, and outer-training methodology.  We include a review of these previous efforts herein.

\subsection{Parameterizations}
Broadly, authors attack the problem of learned optimization at a high level using different parameterizations.  We review these different choices below.

\subsubsection{Controller based parameterization}
One class of learned optimizer parameterizes some function, usually as a neural network, that returns hyperparameters for an existing hand designed method. These methods impose a strong inductive bias. All inner-learning must be done via manipulation of existing methods. By enforcing a constrained structure, these method limit the types of learning rules expressible. This limitation does inject strong priors into the model making both outer optimization easier, as well as produce better outer-generalization. These methods often additionally make use off derived features from the currently training model. These features include things like loss values, variance of function evaluations, and variance of weight matrices.

There are a number of existing works that explore different types of features and architectures. \citet{daniel2016learning} explores using a simple, linear policy which maps from hand designed features to log learning rate which is then used with SGD, RMSProp or momentum.
\citet{xu2017reinforcement} learns a small LSTM network with inputs of current training loss, and predicts a learning rate for SGD.
\citet{xu2019learning} uses an LSTM or an MLP with features computed during training and outputs a scalar which is multiplied by the previous learning rate.

\subsubsection{Symbolic parameterizations}
Existing learning rules are often symbolic in nature. All first order optimizers used today are symbolic and expressible by a small collection of mathematical operations. \citet{Bello17} take inspiration from this and also parameterize optimizers as collections of mathematical primitives. This parameterization can lead to interpretable optimizers. For example, \citet{Bello17} show common subpatterns discovered based on multiplication of the sign of the gradient and sign of the momentum value. Not all computations lend themselves to a symbolic regression. Finding symbolic formula to solve increasingly more difficult tasks starts to resemble program synthesis which is notoriously difficult.

\subsubsection{Continuous parameterizations}
The last family of parameterization is based on continuously parameterized function approximations.
Older work often makes use of simple / linear combinations of features while more recent work makes use of neural networks.
There are a few classes of update rule parameterizations.

\citet{bengio1992optimization} proposes a 7 parameter, and 16 parameter update rule parameterized by mixing various biologically inspired learning signals.
\citet{runarsson2000evolution} use a simple continuous parameterization that operates on error signals and produces changes in inner-parameters.

Recently, there has been renewed interest in learned optimizers, in particular using neural network paramerizations of the learning rules. \citet{andrychowicz2016learning} makes use of an LSTM\citep{hochreiter1997long} possibly with global average pooling to enable concepts like $\ell_2$ gradient clipping to be implemented \citep{bengio2013advances}. They also explore Neural Turing Machine \citep{graves2014neural} like parameterizations specifically designed for low rank memory updates so that it can learn algorithms like LBFGS \citep{nocedal1980updating, liu1989limited}.  In contrast, \citet{metz2019understanding} makes use of a per-parameter MLP instead of the LSTM.

One critical design choice when designing neural network parameterized learned optimizers is input features. When training neural network models it is critical that the inputs be similar scales to aide in optimization. \citet{andrychowicz2016learning} trains on raw gradients which are scaled by either decomposing each scalar into a sign and a magnitude before feeding into an LSTM. \citet{lv2017learning} uses the gradient and momentum normalized by the rolling average of gradients squared (similar to the rescaling done by Adam). \citet{wichrowska2017learned} also makes use of momentum except using multiple timescales normalized in a similar way.
In addition to these momentum terms, \citet{metz2019understanding} other features such as weight value and additionally applies a normalization based on the second moment of features.

In \citep{andrychowicz2016learning, lv2017learning, metz2018learning} all methods employ a per-parameter update. The updates are independent of the number of parameters \footnote{\citep{metz2019understanding} uses normalization across parameters so this is not strictly true}. While expressive, this is expensive as no computation can be shared. \citet{wichrowska2017learned} improve upon this by additionally having per-layer, and a global LSTM.

When designing learned optimizer architectures, there are a number of design decisions to keep in mind. One must balance compute cost of the learned optimizer with expressibility. Often this shifts models to be considerably smaller than those used in supervised learning. \citet{wichrowska2017learned}, for example makes use of a 8-hidden unit LSTM per-parameter. Selection of features to feed into the learned optimizer is also critical. While traditionally deep learning involves learning every feature, this learning comes at the cost of increased compute which is often not feasible.

\subsection{Data distribution of tasks} \label{sec:outer_data}
There is no agreed upon standards when defining datasets for training learned optimizers. The community is adhoc, training on what ever dataset is available or what ever best suites the goals (training a particular model, creating more general optimizers). Constructing large distributions of tasks is labor intensive and thus not often done.

\citet{metz2019understanding} draws inspiration from the few shot learning literature and constructs 10 way classification problems sampling from different classes on imagenet.

\citet{wichrowska2017learned} leverages a large distribution of synthetic tasks. These tasks are designed to represent different types of loss surfaces that might be found in loss surfaces.

TaskSet, \citep{metz2020using} is a dataset of tasks specifically designed for learned optimizer research. We use this dataset throughout our work.

There is a balance between performance of the tasks, and ability to outer-train. Selecting the types of problems we want to train on is often not enough. Additionally, outer-training on the closest task possible will not produce the best optimizer nor even converge. Training on distributions with increased variation smooths the outer-loss surface and makes exploration simpler. This mirrors phenomena found in RL \citep{heess2017emergence, cobbe2018quantifying}.

\subsection{Outer-Optimization methods}
The outer-optimization problem consists of finding a particular set of weights, or configuration of a learned optimizer. A number of different strategies have been proposed.

\subsubsection{Hyper parameter optimization}
One of the most common outer-learning methods used is hyper-parameter optimization. In the context of learning optimizers, this can be seen as finding optimizer hyper parameters, e.g. learning rate, over a particular task instance. Numerous methods exist to do this ranging from Bayesian hyper parameter optimization \citep{shahriari2015taking}, to grid search, to genetic algorithms. See \citet{golovin2017} for a more complete description.

The types of outer-learning problems encountered in learned optimizers is different in that the evaluation function is often an expectation over some distribution. Additionally, the amount of outer-parameters is often larger than simply finding a few hyperparameters. Never the less, we include this here to show the similarity to learned optimizer research.

\subsubsection{Reinforcement learning} \label{sec:outer_opt:rl}
Learned optimizers can naturally be cast into a sequential decision process. The state of the system consists of the inner-parameter values, action space is the steps taken, and the reward is achieving a low loss in the future. A number of works have thus taken this viewpoint.
\citet{li2016learning, li2017learning} makes use of the guided policy search algorithm. \citet{xu2019learning} uses PPO \citet{schulman2017proximal}, \citet{daniel2016learning} make use of Relative Entropy Policy Search \citep{peters2010relative}. The exact algorithm used is a function of the underlying parameterization.

\subsubsection{Neural Architecture Search Style}
Instead of learning the policy directly, \citet{Bello17} makes use of reinforcement learning (PPO \citep{schulman2017proximal} to learn a controller which produces the symbolic learned optimizer. This is distinct from \sref{sec:outer_opt:rl} as it does not leverage the sequential nature of the inner problem. Instead, it treats the environment as a bandit problem.

\subsubsection{Backpropogation / Gradient based}
Gradient based methods leverage local perturbations in parameter space. Computing derivatives through leaning procedures has been explored in the context of hyperparameter optimization in \citep{bengio2000gradient, baydin2014automatic, maclaurin2015gradient}. \citet{andrychowicz2016learning} was the first to make use of gradient based learning for this application.

To train a learned optimizer, ideally, one would compute the derivative of the entire training run with respect to optimizer parameters. This is often referred to as \textit{unrolling} the entire training procedure into one large graph then running reverse mode automatic differentiation on this. Not only is this often too expensive to do in practice, the resulting loss surface can be poorly conditioned \citep{metz2019understanding}. As such approximations are often made.

One common family of approximation is based on truncated backpropogation through time. The core idea is to break apart long unrolled computations into shorter sequences and thus not propagating any error back through the entire unrolled computation graph pieces~\citep{werbos1990backpropagation, tallec2017unbiasing}.  This approximation is used widely in neural network parameterized learned optimizers \citep{andrychowicz2016learning, wichrowska2017learned, lv2017learning, metz2019understanding}. Unlike in language modeling, truncated backprop has been shown to lead to dramatically worse solutions for meta-learning applications \citep{wuunderstanding, metz2019understanding}.

A second family of approximations involve first order gradient calculations. \citet{andrychowicz2016learning} does use the first order gradient calculation where as subsequent work, \citep{wichrowska2017learned} does and computes the full gradient. The trade offs between these two gradient estimators has been discussed in the few shot learning literature in the context of MAML / Reptile \citep{finn2017model, nichol2018first}.

Computing gradients through iterative, non-linear, dynamics has been shown to cause chaotic dynamics. \citet{pearlmutter1996investigation, maclaurin2015gradient} showed high sensitivity to learning rate with respect to performance after multiple steps of unrolled optimization. \citet{metz2019understanding} shows this issue for neural network parameterized learned optimizers and proposes a solution based on variational optimization and multiple gradient estimators.

Despite improvements, there are a lot of considerations that must be taken into account for doing gradient based training. 

\subsubsection{Evolutionary Strategies}
A alternative way to estimate gradients is to use black box method such as Evolutionary Strategies\citep{rastrigin1963convergence, rechenberg1973evolutionsstrategie, nesterov2011random, choromanski2018structured, salimans2017evolution}. These methods are memory efficient, requiring no storage of intermediate states, but can suffer from high variance. In the case of learned optimizer optimization, however, these methods can result in lower variance gradient estimators~\citep{metz2019understanding}. Hybrid approaches that leverage both gradients and ES have been such as Guided ES \citep{maheswaranathan2019guided} have also been proposed for meta-optimization. This work leverages one of the simplest forms of evolutionary strategies as described in ~\citep{salimans2017evolution} which uses a fixed standard deviation.

\section{Outer Optimization Details}
\label{app:outer_opt}
In this work, as with \citet{metz2018learning}, we use asynchronous, batched training. Each task has a different complexity, thus will produce outer-gradient estimates at a different rate. We use asyncronous minibatched training as synchronous training with these heterogenious workloads would be too slow and wasteful. We tie the outer batch size to the number of workers. To prevent stale gradients, we additionally throw away all outer-gradient estimates that are from more than five outer-iterations away from the current weights.

We optimize all models with Adam. We sweep learning rates between 3e-5 and 3e-3 for all experiments. We find the optimal learning rate is very sensitive and changes depending on how long outer-training occurs.
We have preliminary explorations into learning rate schedules but have not yet been able to improve on this constant schedule. For all outer-training experiments, we always run more than one random seed. Due to the relatively small number of units, and biased gradient estimators, performance is dependant on random seed. For all experiments we use gradient clipping of 0.1 applied to each weight independently. Without this clipping no training occurs. This surprises us as our gradient estimator is evolutionary strategies which will not typically have exploding gradients. Upon further investigation, however, the outer-gradient variance is much larger without this clipping.

When computing outer-gradients, we follow~\citep{metz2019understanding} and compute a outer-loss based on multiple mini batches of data. In our work we use 5. Note inner-training always uses a single minibatch of inner-training data as well as a single batch of inner-validation data when used.

We first train with 240-360 length unrolls over a max of 10k inner-steps. While training we logged out 10k length unrolls from 100 tasks sampled from the outer-training distribution and saved outer-parameters every hour.
While training we monitor performance across all outer-learning rates and all seeds on the outer-training distribution. When training plateaus, we manually look through these evaluations and select a candidate set of optimizers to further train with an increasing truncation schedule. Not all optimizers fine tune in the same way despite having the same performance on the outer-training data so selecting more than one is critical. At this point we are unsure where this phenomenon comes from.

Next we fine tune theses models in an unbiased fashion. We explored two methods. First, based on an increasing truncation schedule. We tested linearly increasing truncation length from 300-10k steps over the course of 30k or 10k steps. We find the faster increase, 10k steps, performs best. Second, we explored fine tuning with Persisted Evolutionary Strategies -- an unbiased gradient estimator~\citep{pes}. We found this achieved similar final performance but achieved it in half the time.
When fine tuning we also make use of different learning rates. We find higher learning rates make progress faster, but can be unstable in that the performance varies as a function of outer-training step. Additionally, the learning rate chosen depends on the learning rate used previously in the first training phase. Before finetuning, we `warm up' the Adam internal rolling statistics. While this might not be strictly required, it ensures that there is no decrease in performance early in outer-training. This can be done by simply setting the outer-learning rate to zero for the first 300 outer-iterations.

\section{Learned Optimizer Architecture Details}
\label{app:learned_opt_details}
In this section we describe the detailed learned optimizer architecture. For ease of understanding we opt to show a mix of pseudo-code based on python and textual descriptions as opposed to mathematical expressions. Finally we chose to describe our optimizer as a series of stateless and pure functions for clarity. 

We used this architecture for all of our experiments except of Fig~\ref{fig:meta_train}b which used an older version of the architecture with additional features which where dropped from the final version.

\subsection{High level structure of the optimizer}
\label{details:learner_state}
The learned optimizer has two main components: a function that maps from some set of inputs, a state, and parameters to some new state and new parameters.
\begin{python}
class Optimizer:
    def next_state(inputs: Inputs,
                   state: State,
                   parameters: Params
                   ) -> (State, Params);
\end{python}
And a function to produce an initial state from the given inner-parameters.
\begin{python}
    def initial_state(self, params: Params) -> State;
\end{python}

Parameters are stored as a dictionary of different tensors keyed by name. 
\begin{python}
Params = Dict[Text, Tensor]
\end{python}

For convenience, we also define gradients to be the same type as Params:
\begin{python}
Grads = Params = Dict[Text, Tensor]
\end{python}

The state consists of multiple values that we will discuss in detail further. For now, however, we list the full state with high level comments.
\begin{python}
State = namedtuple(
    "State",
    [
        # current inner training iteration
        "training_step" : Int 
         # the statistics for the rolling averages. 
        "rolling_features": RollingFeatureState,
        # Hidden state of the lstm
        "lstm_hidden_state": LSTMHiddenState, 
         # Activations passed from the MLP to the LSTM.
        "from_mlp": List[Tensor[shape=(from_mlp_size,)]],
        # Activations from LSTM passed to LSTM
        "from_lstm": Tensor[shape=(from_lstm_size,)], 
        # Rolling statistics of the train loss value
        "train_loss_accum": LossAccumState, 
         # Rolling statistics of the valid loss value
        "valid_loss_accum": LossAccumState,
         # State to manage inner-gradinet clipping.
        "dynamic_clip": RollingClipState,
         # Parameter values from the nearest 100 steps in the past.
    ])
\end{python}
Both from\_lstm\_size, from\_mlp\_size are hyper parameters set as part of the learned optimizer to control how much information is sent from the MLP or from the LSTM.

The input to the next\_state function consists of inner-gradients, computed on a inner-training batch of data, as well as optionally validation data which is passed in every 10 iterations. We choose to not pass in validation data every steps for computational efficiency.
\begin{python}
Inputs = namedtuple(
    "Inputs",
    [
        # gradient from task. This is same type as the parameters.
        "inner_grads": Grads, 
        # training loss computed from a mini batch.
        "train_loss": float, 
        # validation loss from a mini batch of validation data.
        "valid_loss": Optional[float],
    ])
\end{python}

Each task specifies a function that samples parameter initialization, as well as to produce outputs.
\begin{python}
class Task:
    def initial_params(self) -> Params
    def inputs_for_params(self, params) -> (
                Grads, # inner-training gradients
                float, # inner-training loss
                Optional[float]) # optional inner-training validation
\end{python}

For outer-training each task also includes a second loss function which computes the task's loss on the outer-validation split of data. Note this uses a \emph{different} validation set of data than the previous function.
\begin{python}
    def valid_outer_loss(self, params: Params) -> float
\end{python}

Inner training / application of the learned optimizer looks like:
\begin{python}
task = SampleTask()
optimizer = Optimizer()

params = task.initialParams()
state = optimizer.initial_state(state)

for i in range(N): # Length of inner training
    inputs = GetInputFromTask(params, with_valid_data=(i % 10 == 0))
    params, state = optimizer.next_state(inputs, params, state)
\end{python}

Computing the outer objective and outer-gradients from inner-initialization looks like the following:
\begin{python}
task = SampleTask()
optimizer = Optimizer()
# When outer-training with ES, perturb optimizer weights.

params = task.initialParams()
state = optimizer.initial_state(state)

outer_valid_losses = []
for i in range(N): # Length of a trunction
    inputs = GetInputFromTask(params, with_valid_data=(i % 10 == 0))
    params, state = optimizer.next_state(inputs, params, state)
    for i in range(10):
        outer_valid_losses.append(task.valid_outer_loss(params))

outer_loss = mean(outer_valid_losses)
# When outer-training with ES, one must perturb the optimizer's parameters before computing outer_loss.
\end{python}

\subsection{Utilities / components}

First we will describe the individual components and utilities used, then we will go on to describe the full update function.

\subsubsection{Rolling Features}

These are a moving average of gradients and second moments computed similarly to Adam / RMSProp. The rolling state consists of tensors containing momentum values (ms) and second moment values (rms):

\begin{python}
RollingFeaturesState = collections.namedtuple("RollingFeaturesState",
                                    ["ms": Dict[Text, Tensor],
                                    "rms": Dict[Text, Tensor]])
\end{python}

The Text represent names that map to the tensor of the corresponding shape from the parameters.
The values are the same shape as the corresponding inner-parameter with an additional axis appended to keep track of multiple different decay values. While its possible to outer-learn these values, we fix them at 0.5, 0.9, 0.99, 0.999, 0.9999.

To update these we construct a helper class.
\begin{python}
Class RollingState:
    def __init__(self, decay_values):
        self.decay_values = decay_values
\end{python}

We define an initial value which is simply simply zeros:
\begin{python}
    def initial_state(self, params: Params): -> RollingFeatureState
        n_dims = len(decay_values)
        ms = {k: tf.zeros([v.shape + [n_dims]}
        rms = {k: tf.zeros([v.shape + [n_dims]}
        return RollingFeaturesState(ms=ms, rms=rms)
\end{python}

To update these values we follow a procedure similar to RMSProp
update equations:
\begin{python}
    def update_state(self, state: RollingFeatureState, grads: Grads) -> RollingFeatureState:
        ret_ms = {}
        ret_rms = {}
        for k in state.keys():
            g = grads[k]
            s = state[k]
            
            ret_ms[k] = np.zeros_like(ret.ms[k])
            ret_rms[k] = np.zeros_like(state.rms[k])

            for di, decay in enumerate(self.decay_values):
                ret_ms[k][..., di] = ret[k][..., di] * decay + (1 - decay) * grad
                ret_rms[k][..., di] = ret[k][..., di] * decay + (1 - decay) * grad**2
        return RollingFeatureState(ms=ret_ms, rms=ret_rms)
\end{python}

\subsubsection{LossAccum}
This represents how we get loss information into our learned optimizer. Loss values have no pre-determined scale and span many orders of magnitude.
As such, we must somehow standardize them so that the inputs are bounded and able to be easily used by neural networks.
We get around this by keeping track of the normalized mean and variance of the mini-batch losses.

\begin{python}
LossAccumState = collections.namedtuple("LossAccumState",
                ["mean": float, # rolling mean loss value
                 "var": float, # rolling second moment loss value
                 "updates": int # Number of updates performed so far
                ])
\end{python}

The class that manages these states is parameterized by the decay of the rolling window.

\begin{python}
class LossAccum:
  def __init__(self, decay):
    self.decay = decay
\end{python}

The initial state is simply zeros.

\begin{python}
  def initial_state(self) -> LossAccumState:
    return LossAccumState(
        mean=tf.constant(0., dtype=tf.float32),
        var=tf.constant(0., dtype=tf.float32),
        updates=tf.constant(0, dtype=tf.int64))
\end{python}

To compute updates we do rolling mean and variance computations.
\begin{python}
  def next_state(self, state: RollingAccumState, loss: float): -> LossAccumState
    new_mean = self.decay * state.mean + (1.0 - self.decay) * loss
    new_var = self.decay * state.var + (
        1.0 - self.decay) * tf.square(new_mean - loss)
    new_updates = state.updates + 1
    return LossAccumState(mean=new_mean, var=new_var, updates=new_updates)
\end{python}

Two additional functions are used to normalize loss values for use in neural networks. First, we have a ``corrected'' mean (similar to what is done by Adam) for a given AccumState.
\begin{python}
  def corrected_mean(self, state: AccumState) -> float:
    c = 1. / (1 - self.decay**tf.to_float(state.updates) + 1e-8)
    return state.mean * c
\end{python}

Second, we have a function that weights one loss by a different AccumState.
This is eventually used to weight the validation loss accum against the training loss state allowing the learned optimizer to detect overfitting.

\begin{python}
  def weight_loss(self, state: AccumState, loss: float) -> float:
    c = 1. / (1 - self.decay**tf.to_float(state.updates) + 1e-8)
    cor_mean = state.mean * c
    cor_var = state.var * c
    l = (loss - cor_mean) * tf.rsqrt(cor_var + 1e-8)
    return tf.clip_by_value(l, -5, 5)
\end{python}

\subsubsection{RollingClipState}
Gradient clipping is a often used technique in deep learning. We found large benefits by applying some form of learned clipping (clipping inner-gradient values) in our learned optimizer.
We cannot simply select a default value because gradient norms vary across problem.
As such we meta-learn pieces of a dynamic gradient clipping algorithm.
This is our first iteration of this concept and we expect large gains can be obtained with a better scheme.

This algorithm is stateful thus also needs some state container.

\begin{python}
RollingClipState = (int, # the number of times this has been updated,
                    float # rolling average of mean of squared gradient values.
                    )
\end{python}

This class is parameterized by the decay constant of the rolling average and the multiplier to determine when clipping should start. Both of these values are outer-learned with the rest of the learned optimizer.

\begin{python}
class RollingGradClip:
  def __init__(self, alpha=0.99, clip_mult=10):
    self.alpha=alpha
    self.clip_mult=clip_mult
\end{python}

The initial states are initialized to 1.
\begin{python}
  def initial_state(self) -> RollingClipState:
    return (tf.constant(1, dtype=tf.float32),
            tf.constant(1.0, dtype=tf.float32)*(1-self.alpha))
\end{python}

We provide a normalization function that both updates the RollingClipState and provides clipped gradients.
\begin{python}
  def next_state_and_normalize(self,
                    state: RollingClipState,
                    grads: Grads) ->
                        RollingClipState, Grads:
    def _normalize(state: RollingClipState, grads: Grads):
        summary_ops = []
        t, snd = state
        clip_amount = (snd / (1-self.alpha**t))*self.clip_mult
        clipgs = [tf.clip_by_value(g, -clip_amount, clip_amount) for g in grads.values()]
        return clipgs
  
    t, snd = state
    clipped_grads = self._normalize(state, grads)
    mean_square_list = [tf.reduce_mean(tf.square(g)) for g in clipped_grads]
    new_snd_moment = tf.sqrt(1e-8 + tf.reduce_mean(mean_square_list))
    next_snd = snd * self.alpha + new_snd_moment * (1. - self.alpha)
    return (t+1, next_snd), clipped_grads
\end{python}

\subsection{Learned optimizer specific / putting it all together}
The learned optimizer has two main components. The first is the Optimizer class that manages everything surrounding inner-learning. This function has no learnable outer-parameters.
The second is what we call ``theta\_mod'' which contains all of the outer-variables and functions that we are learning.

First we construct the optimizer class with the corresponding theta\_mod (see bellow). 
\begin{python}
class Optimizer:
  def __init__(self,theta_mod):
    self.theta_mod = theta_mod
\end{python}

Next the rolling features at fixed, hard coded intervals. These could be outer-learned but in this work they are fixed.
\begin{python}
    self.rolling_features = RollingFeatures(
      decays=[0.5, 0.9, 0.99, 0.999, 0.9999])
\end{python}

Then the the loss features for use with both training and validation loss. We use a lower decay constant on the validation as it is updated less frequently (once every 10 steps). Once again these could be outer-learned but in this work we leave them fixed.
\begin{python}
    self.train_loss_accum = LossAccum(0.95)
    self.valid_loss_accum = LossAccum(0.9)
\end{python}

Finally we construct the gradient clipping utility. Unlike the previous parameters, we do outer-learn these and pass in two variables off of the `theta\_mod'.
\begin{python}
    self.rolling_clip = RollingGradClip(
                        self.theta_mod.rolling_clip_alpha,
                        self.theta_mod.rolling_clip_mult)
\end{python}

Next, we need an initial state for the optimizer defined in \ref{details:learner_state}. This mostly consists of obtaining initial states from the various components of the learned optimizer.

\begin{python}
  def initialState(self, params : Params) -> State:
    shapes = [v.shape.as_list() for v in params.values()]
    return State(
        training_step=0,
        rolling_features=self.rolling_features.initial_state(shapes),
        from_lstm=self.theta_mod.initial_from_lstm(),
        from_mlp=self.theta_mod.initial_from_mlp(len(shapes)),
        lstm_hidden_state=self.theta_mod.initial_rnn_state(len(shapes)),
        train_loss_accum=self.train_loss_accum.initial_state(),
        valid_loss_accum=self.valid_loss_accum.initial_state(),
        dynamic_clip=self.rolling_clip.initial_state(),
    )
\end{python}

Next we look to the update performed. We split up into two components -- first a validation state, and second a training state.

\begin{python}
  def next_params_and_state(self, params: Params,
                            current_state: State,
                            train_loss: float,
                            train_grads: Grads,
                            valid_loss: Optional[float]) -> (Params, State):
    if valid_loss:
        current_state = self.next_state_validation(current_state, valid_loss)
        
    params, next_state = self.next_state_trainining(params, current_state, train_loss, train_grads)
    return params, next_state
\end{python}

The validation update simply consists of updating the two components that make use of the loss value.
\begin{python}
  def next_state_validation(self, current_state: State,
                            valid_loss: float) -> State:
    next_valid_loss_accum = self.valid_loss_accum.next_state(
        current_state.valid_loss_accum, valid_loss)
    return current_state._replace(
                    valid_loss_accum=next_valid_loss_accum)
\end{python}

The the function applied to the training gradients is much more complicated. Much like the validation features it updates the various rolling statistics including the two loss data structures, the rolling momentum / rms terms, as well as the gradient clipping state.
\begin{python}
  def next_state_trainining(self,
                params: Params,
                current_state: State,
                loss: float,
                grads: Grads) -> (Params, State):
    grads = list(grads.values())
    next_train_loss_accum = self.train_loss_accum.next_state(
        current_state.train_loss_accum, loss)
    next_dynamic_clip, grads = self.rolling_clip.next_state_and_normalize(current_state.dynamic_clip, grads)
    next_rolling_features = self.rolling_features.next_state(
        current_state.rolling_features, grads)
\end{python}

Next, a sequence of features related to loss values are computed.
\begin{python}
    train_loss_feat = self.train_loss_accum.weight_loss(next_train_loss_accum, loss)
    valid_loss = self.valid_loss_accum.corrected_mean(current_state.valid_loss_accum)
    valid_loss_feat = self.train_loss_accum.weight_loss(next_train_loss_accum, valid_loss)
\end{python}

Next we call into the learned optimizer function in ``theta\_mod'' which does the bulk of the computation.
\begin{python}
    next_params, from_lstm, from_mlp, lstm_hidden_state = \
      self.theta_mod.compute_update_and_next_state(
            train_loss_feat, valid_loss_feat,
            current_state.from_mlp,
            current_state.from_lstm,
            current_state.lstm_hidden_state,
            next_rolling_features,
            grads,
            params,
            current_state.training_step,
            )    
\end{python}

We then populate the next state with all the outputs.
\begin{python}
     next_state = State(
        rolling_features=next_rolling_features,
        training_step=current_state.training_step + 1,
        lstm_hidden_state=lstm_hidden_state,
        from_lstm=from_lstm,
        from_mlp=from_mlp,
        train_loss_accum=next_train_loss_accum,
        valid_loss_accum=current_state.valid_loss_accum,
        dynamic_clip=next_dynamic_clip,
    )   
    
    return next_params, next_state
\end{python}

\subsection{``ThetaMod'': The learned optimizer outer-parameters}
All of the meta-learned outer-parameters are stored in a class called ThetaMod.
Its constructor defines various dimensions of the different Sonnet modules.

\begin{python}
class ThetaMod:

  def __init__(self,
              # Size of the features from the per parameter mlp.             
              from_mlp_size=16, 
              # size of the features from the per tensor LSTM.
              from_lstm_size=18,
              # Size of features from the lstm being passed to the per param MLP.
              lstm_to_ff=17, 
              # LSTM hidden size.
              lstm_hidden_size=64,
              # Multiplier on output to improve conditioning
              step_multiplier=0.001,
              # Multiplier on output to improve conditioning
              magnitude_rate=0.001,
              **kwargs):
    self.step_multiplier = step_multiplier
    self.magnitude_rate = magnitude_rate
\end{python} 
This constructor creates a shared sonnet module for the per layer LSTM. This is constructed here as it is needed in more than one method.
\begin{python}
      self.rnn = InitialStateLSTM(lstm_hidden_size)
\end{python}
Second, we construct the per parameter feed forward network. This is a tiny network that operates on each parameter. It outputs two values for producing a step, then from\_mlp\_size more values which are fed back into the LSTM.
\begin{python}
      self.ffmod = snt.nets.MLP([32, 32] + [2 + self.from_mlp_size], name="PerParamMLP")
\end{python}
Third, we construct two linear projections that map from the RNN output to either the per parameter feed forward network, or back to the global features.
\begin{python}
      self.rnn_to_ff = snt.Linear(lstm_to_ff, name="rnn_to_ff")
      self.rnn_to_in = snt.Linear(from_lstm_size, name="rnn_to_in")
\end{python}

Finally, we initialize the two variables used by the rolling gradient clipping terms. We parameterize these variables on a log scale (possibly subtracting 1) so as to roughly match the scaling of the other neural network weights.

\begin{python}
      def get_log_param_variable(name,
                                 initial_value,
                                 minus_one=False):
        if minus_one:
          initial_value = 1 - initial_value
        init_log = np.log(initial_value)
        v = tf.get_variable(
            name,
            initializer=tf.ones(shape=[]) * tf.constant(
                init_log, dtype=tf.float32),
                trainable=True)
        ret = tf.exp(v)
        if minus_one:
          ret = 1 - ret
        return ret

      self.rolling_clip_alpha = get_log_param_variable(
                name="rolling_clip_alpha",
                initial_value=0.99,
                minus_one=True)

      self.rolling_clip_mult = get_log_param_variable(
                name="rolling_clip_mult",
                initial_value=10.0,
                minus_one=False)
\end{python}

The heart of this function takes the features passed in from the optimizer, applies some neural network based learning algorithm, then produces the next weight value as well as various pieces of hidden information for the next iteration.
\begin{python}
  def compute_update_and_next_state(self,
        # Training loss feature
        train_loss_feat: Tensor[shape=()],
        # Validation loss feature
        valid_loss_feat: Tensor[shape=()], 
        # Information passed to each layer of the LSTM from the per param MLP.
        from_mlp: List[Tensor[shape=(from_mlp_size,)]], 
        # Information aggregated across LSTM layers to be passed to all LSTM layers.
        from_lstm: Tensor[shape=(from_lstm_size,)],
        # The current LSTM state.
        lstm_state: LSTMHiddenState, 
        # Features from the momentum / rolling second moments.
        rolling_features: RollingFeatureState, 
        # Gradients from current training mini-batch
        grads: Grads, 
        # Current parameters
        params: Params, 
        # current inner-iteration
        training_step: int, 
        ) ->
            # the output parameter value.
            (Params, 
            # next from_lstm
            Tensor[shape=(from_lstm_size,)],
            # next from_mlp
            List[Tensor[shape=(from_mlp_size,)]], 
            # next lstm hidden state.
            LSTMHiddenState): 
\end{python}

We first compute the global feature vector. These consist of features related to the loss curves, as well as the aggregated data from the previous LSTM execution.
These features form a rank 1 tensor.
\begin{python}
    global_features = self.compute_global_features(train_loss_feat,
                                                   valid_loss_feat,
                                                   from_lstm)
\end{python}

Next we compute the per tensor features. This is a rank two tensor, consisting of the number of tensors, and a feature dimension.
\begin{python}
    per_tensor_features  = self.compute_tensor_features(grads,
                                                        params,
                                                        from_mlp,
                                                        rolling_features)
\end{python}

We tile the global features to be num tensors by num global features then concat these to the per tensor features and pass them into the per layer LSTM.
\begin{python}
    rnn_inputs = tf.concatenate([per_tensor_features,
            tf.tile(global_features, [per_tensor_features.shape[0], 1])])
    lstm_out, next_lstm_hidden_state = self.rnn(rnn_inputs, lstm_state)
\end{python}

We then linearly project the LSTM outputs to be input to the per parameter feed foward network, and unstack them to form a a list of rank 1 features -- one per tensor.

\begin{python}
    ff_inputs = self.rnn_to_ff(lstm_out)
    ff_inputs = tf.unstack(ff_inputs)
\end{python}

Next we iterate over each parameter, and first compute per parameter features.
The per parameter features consist of gradients, parameter values,
parameter values 100 steps in the past, momentum values, second moment values, activations from the per tensor LSTM.
Additionally there are two global features included: the training step and the number of tensors total. Two empty lists are initialized to accumulate results.
\begin{python}
    next_to_lstms = []
    next_params = []
    
    for wi in range(len(params)):
        feats = compute_per_param_features(grads[wi],
                                          param[wi],
                                          rolling_features.ms[wi],
                                          rolling_features.rms[wi],
                                          ff_inputs[wi]
                                          training_step,
                                          len(params))
\end{python}
We pass these per parameter features through these features into a per parameter MLP which produces a single output.

\begin{python}
        output = self.ffmod(feats)
\end{python}

We split this output into three pieces: A direction, a magnitude, and features which are  be fed back to the per tensor LSTM after being reduced along the parameter dimension.
\begin{python}
        direction = output[:, 0:1]
        magnitude = output[:, 1:2]
        to_lstm = tf.reduce_mean(output[:, 2:], axis=0)
\end{python}

A step can then be computed by exponentiating the magnitude and multiplying by a direction.
Rescaling these quantities is critical. Without this rescaling, the magnitude of the updates is extremely large, resulting in chaotic training dynamics which makes outer-learning difficult.

\begin{python}
        step = direction * tf.exp(magnitude * self.magnitude_rate) * self.step_multiplier
\end{python}

We then subtract step from the previous parameter value. While it is possible just to predict a new parameter directly, learning deltas is significantly easier to learn and results in much more stable training.
\begin{python}
        next_param = param[wi] - step
\end{python}

Finally, we collect all the information needed to be returned from the per parameter LSTM and output.
\begin{python}
        next_params.append(next_param)
        next_to_lstms.append(next_to_lstm)
\end{python}

With the for loop over, we then return the necessary info. Before this, however, we must aggregate information from the per tensor LSTM to be able to pass it to the next layer's global features after applying a linear projection from the LSTM output and reducing along the tensor dimension.

\begin{python}
    from_lstm = tf.reduce_mean(self.rnn_to_in(lstm_out), axis=0)
    return next_params, from_lstm, next_to_lstms, next_lstm_hidden_state
\end{python}

Now we will discuss the different features methods. First, computing the global features.
\begin{python}
    def compute_global_features(self,
                            train_loss_feat: Tensor[shape=()],
                            valid_loss_feat: Tensor[shape=()],
                            from_lstm: Tensor[shape=(n_lstm_out,)]
                            ): -> Tensor[shape=(n_global_features,)]
\end{python}

This function simply returns the concatenation of the Two loss features, and the aggregated features from the per layer lstm.
\begin{python}
    return tf.concatenate([train_loss_feat, valid_loss_feat], from_lstm])
\end{python}

Next we look to the per tensor features. We return a 2D tensor with the leading dimension a ``batch size'' being the number of tensors, and the second dimension representing the feature dimension.

\begin{python}
    def compute_tensor_features(self,
            grads: Grads,
            params: Params,
            from_mlp: List[Tensor[shape=(from_mlp_size,)]],
            rolling_features: RollingFeatureState
            ): -> Tensor[shape=(n_tensors, n_features)]
\end{python}

We compute a large number of features. At this point have not extensively ablated each one and we expect many are duplicate or unneeded. Before we begin listing features, however, we make use of a helper function that computes the clipped log abs of tensors.

\begin{python}
        def clip_log_abs(v):
          mag = tf.log(1e-8 + tf.abs(v))
          return tf.clip_by_value(mag, -5, 5)
\end{python}

We begin by iterating over each tensor.

\begin{python}
    stacked_inputs = []
    for i in range(len(params)):
      inputs = {}
\end{python}

First, we compute features based on the rolling momentum values. We compute features that represent the log magnitude, the sign, and the log variance.
\begin{python}
      mean_ms = tf.reduce_mean(rolling_features.ms[i])
      inputs["mean_ms_mag"] = clip_log_abs(mean_ms)
      inputs["mean_ms_sign"] = tf.sign(mean_ms)
      var_ms = tf.reduce_mean(tf.square(rolling_features.ms[i] - mean_ms))
      inputs["var_ms"] = clip_log_abs(var_ms)
\end{python}

We compute a similar set of features for the second moment features.
\begin{python}
      mean_rms = tf.reduce_mean(rolling_features.rms[i])
      inputs["mean_rms"] = clip_log_abs(mean_rms)
      inputs["mean_sign"] = tf.sign(mean_rms)
      var_rms = tf.reduce_mean(tf.square(rolling_features.rms[i] - mean_rms))
      inputs["var_rms"] = clip_log_abs(var_rms)
\end{python}

A similar set for the inner-parameters:
\begin{python}
      v = params[i]
      mean_v = tf.reduce_mean(v)
      inputs["mean_v_mag"] = clip_log_abs(mean_v)
      inputs["mean_v_sign"] = tf.sign(mean_v)
      var_v = tf.reduce_mean(tf.square(v - mean_v))
      inputs["var_v"] = clip_log_abs(var_v)
\end{python}

Next we include the magnitude of the weight norm.
\begin{python}
      inputs["norm_weight"] = clip_log_abs(tf.norm(params[i]))
\end{python}

As well as the gradient norm:
\begin{python}
      inputs["g_norm"] = clip_log_abs(tf.norm(grads[i]))
\end{python}

Next we add a feature if the tensor is a scalar or not.
\begin{python}
      if len(v.shape.as_list()) == 0:
        is_scalar = tf.constant(1.0)
      else:
        is_scalar = tf.constant(-1.0)
      inputs["is_scalar"] = is_scalar
\end{python}

The shape of the underlying tensor is also a useful feature.
The length of this shape varies though depending on the rank.
As such we first pad the tensor to be at least rank 4 and then pass the log shape shifted by -1.
This shift is to keep this value roughly zero mean.

\begin{python}
      extra_dims = [1.] * (4 - len(v.shape.as_list()))
      shape_stack = tf.concat([tf.to_float(tf.shape(v)),
                              tf.stack(extra_dims)],
                              axis=0)

      for j in range(4):
        # shift so 1D is neg one to be better scaled.
        inputs["shape_%d" % j] = tf.log(shape_stack)[j] - 1.0 
\end{python}

The features from the previous execution's MLP are included. These are to enable communication back from the MLP to the per layer network.

\begin{python}
      inputs["from_mlp"] = from_mlp[i]

\end{python}

This dictionary of features is then flattened (adding appropriate dimensions if needed) and appended to the list of inputs.
\begin{python}
      values = sorted_values(inputs)
      reshaped = [
          tf.expand_dims(v, 0) if len(v.shape.as_list()) == 0 else v
          for v in inputs.values()
      ]
      stacked_inputs.append(reshaped)
\end{python}

Once the stacked\_inputs is filled, we zip and stack and concat the values forming a number of layers by number of features tensor.
\begin{python}
    stacked_features = [tf.stack(v) for v in zip(*stacked_inputs)]
    return tf.concat(stacked_features, axis=1)
\end{python}

The last set of features is the per parameter features.
\begin{python}
    def compute_per_param_features(self,
                grad: Tensor,
                param: Tensor,
                ms: Tensor,
                rms: Tensor,
                ff_inputs: Tensor[shape=(num_ff_inputs,)],
                training_step: int,
                num_params: int
                ): -> Tensor[shape=(num_params, num_features)]
\end{python}

Many of these inputs are of the shape of the underlying tensor. Because many of these operations are per parameter we construct a flat version of all of these.

\begin{python}
        flat_g = tf.reshape(grad, [-1, 1])
        flat_v = tf.reshape(param, [-1, 1])
        flat_ms = tf.reshape(ms, [-1, ms.shape[-1])
        flat_rms = tf.reshape(rms, [-1, rms.shape[-1])
\end{python}

As with the per layer features, we add a series of features. First we add raw features for the gradients, parameter values, momentum, and second moment accumulators, and past weights.
\begin{python}
        inps = {}
        inps["flat_g"] = flat_g
        inps["flat_v"] = flat_v
        inps["m"] = m
        inps["rms"] = rms
\end{python}

We then add some derived features such as the log abs parameter value:
\begin{python}
        inps["log_abs_v"] = tf.log(tf.abs(flat_v) + 1e-8)
\end{python}

Finally we add terms based on the rsqrt of the second moment terms inspired by how Adam normalizes.
\begin{python}
        rsqrt = tf.rsqrt(rms + 1e-6)
        rms_scaled_g = m * rsqrt
        inps["rms_scaled_g"] = rms_scaled_g
        inps["rsqrt"] = rsqrt
\end{python}

These values are concatenated and normalized by a second moment normalizer computed across the num tensor dimension. We additionally clip the output of this normalizer. This is crucial for stable training as some of these features will produce large magnitudes in some channels.

\begin{python}
        inp = tf.concat(sorted_values(inps), 1)
        inp = inp * tf.rsqrt(1e-8 +
                          tf.reduce_mean(tf.square(inp), axis=0, keep_dims=True))
        inp = tf.clip_by_value(inp * 0.5, -1, 1)
\end{python}

Next, we embed the current inner-training step with sinusoids of different frequencies.

\begin{python}
        def sin_embedding(x):
          mix_proj = []
          for i in [1, 3, 10, 30, 100, 300, 1000, 3000, 10000, 30000, 100000]:
            s = tf.to_float(tf.to_float(i) / float(np.pi))
            mix_proj.append(tf.sin(s * tf.to_float(x)))
          return tf.stack(mix_proj)
          
        step = utils.sin_embedding(training_step)
        stack_step = tf.tile(
            tf.reshape(step, [1, -1]),
            tf.stack([tf.shape(flat_g)[0], 1]))
\end{python}

We compute a features based on the number of tensors:
\begin{python}
        log_num_tensors = tf.log(float(len(grads_and_vars))) - 1.

        stack_num_tensors = tf.tile(
            tf.reshape(log_num_tensors, [1, 1]),
            tf.stack([tf.shape(flat_g)[0], 1]))        
\end{python}

As well as features about the number of parameters in the current tensor:
\begin{python}
        log_n_weight = tf.log(tf.to_float(tf.shape(flat_v)[0]))
        stack_log_n_weight = tf.tile(
            tf.reshape(log_n_weight, [1, 1]),
            tf.stack([tf.shape(flat_g)[0], 1]))
\end{python}

Next, we compute some statistics about the log norm of the weight matrix:
\begin{python}
        log_norm = tf.log(tf.norm(flat_v) + 1e-8)
        stack_log_norm = tf.tile(
            tf.reshape(log_norm, [1, 1]),
            tf.stack([tf.shape(flat_g)[0], 1]))
\end{python}

Next, we tile the input from the LSTM so as to add a number of parameters dimension.
\begin{python}
        ff_inp = tf.tile(
            tf.reshape(ff_inputs[i], [1, -1]),
            tf.stack([tf.shape(flat_g)[0], 1]))
\end{python}

Finally we concat and return all these features.
\begin{python}
        return tf.concat([
            inp, stack_step, stack_num_tensors, stack_log_norm,
            stack_log_n_weight, ff_inp
        ], axis=1)
\end{python}

We have discussed how new values of LSTM hidden state, activations from the LSTMs, and activations from the MLP are produced, but we have not yet shown the initial values. In all cases they are initialized from outer-parameterized values shown bellow.
\begin{python}
  @snt.reuse_variables
  def initial_rnn_state(self, n):
    s = self.rnn.initial_state(n, trainable=True)
    return s

  @snt.reuse_variables
  def initial_from_lstm(self):
    return tf.get_variable(
        name="initial_from_lstm",
        shape=[self.from_lstm_size],
        dtype=tf.float32,
        trainable=True)

  @snt.reuse_variables
  def initial_from_mlp(self, n):
    s = tf.get_variable(
        name="initial_from_mlp",
        shape=[self.from_mlp_size],
        dtype=tf.float32,
        trainable=True)
    return [s for _ in range(n)]
\end{python}

\section{Adam8p}
\label{app:adam8p}

This matches the adam8p optimizer described in \citep{metz2020using}. The eight hyper-parameters are: the learning rate, $\alpha$, first and second moment momentum, $\beta_1$, $\beta_2$, the numerical stability term, $\epsilon$, $\ell_2$ and $\ell_1$ regularization strength, and learning rate schedule constants $\lambda_{\text{exp\_decay}}$ and $\lambda_{\text{linear\_decay}}$.

\begin{align}
    \phi^{(0)} =& \text{problem specified random initialization}\\
    m^{(0)} =& 0 \\
    v^{(0)} =& 0 \\
    g^{(t)} =& \frac{d}{d\phi^{(t)}}(f(x; \phi^{(t)}) + \ell_2||\phi^{(t)}||^2_2 + \ell_1||\phi^{(t)}||_1) \\
    m^{(t)} =& \beta_1 m^{(t-1)} + g^{(t)}(1 - \beta_1) \\
    v^{(t)} =& \beta_2 v^{(t-1)} + (g^{(t)})^2(1 - \beta_2) \\
    \hat{m}^{(t)} =& \dfrac{m^{(t)}}{1-\beta_1^{t+1}} \\
    \hat{v}^{(t)} =& \frac{v^{(t)}}{1-\beta_2^{t+1}} \\
    u^{(t)} =& \dfrac{\hat{m}^{(t)}}{\sqrt{\hat{v}^{(t)}} + \epsilon} \\
    s_{\text{linear}}^{(t)} =& \text{max}(1 - t \lambda_{\text{linear\_decay}}, 0) \\ 
    s_{\text{exp}}^{(t)} =& \text{exp}(-t \lambda_{\text{exp\_decay}}) \\
    \phi^{(t+1)} =& \alpha  s_{\text{linear}}^{(t)} s_{\text{exp}}^{(t)} u^{(t)}
\end{align}

We sample learning rate logritmically between 1e-8 and 10, beta1 and beta2 we parametrize as $1-x$ and sample logrithmically between 1e-4 and 1 and 1e-6 and 1 respectively. For learning rate schedules we sample linear decay between  1e-7, 1e-4 logrithmically and exponential decay logrithmically between  1e-3, 1e-6. We sample both $\ell_1$ and $\ell_2$ logrithmcally between 1e-8, 1e1.

\section{Performance Table}
We show numerical performance measurements of our learned optimizer and baseline optimizers. Each value is the mean or median over 100 different inner-training tasks.

\begin{center}
 \begin{tabular}{||c c c c c||} 
 \hline
  & train &  & test &  \\
 Optimizer & mean  & median & mean & median \\ [0.5ex] 
 \hline\hline
learned & 0.077325 & 0.024167 & 0.113242 & 0.028928 \\
\hline
global adam lr (1 Trial) & 0.226807 & 0.105513 & 0.261613 & 0.092532 \\
\hline
global adam8p, RS (1 Trial) & 0.240924 & 0.109519 & 0.257368 & 0.100959 \\
\hline
global nadamw, RS (1 Trial) & 0.215176 & 0.098448 & 0.229359 & 0.086703 \\
\hline
per task adam lr (14 Trial) & 0.150443 & 0.067373 & 0.152028 & 0.063194 \\
\hline
per task opt\_list (10 Trial) & 0.082756 & 0.039647 & 0.086195 & 0.038541 \\
\hline
per task opt\_list (100 Trial) & 0.041560 & 0.028851 & 0.047224 & 0.031042 \\
\hline
per task opt\_list (1k Trial) & 0.039348 & 0.028254 & 0.043193 & 0.029897 \\
\hline
per task adam8p (10 Trial) & 0.278858 & 0.132965 & 0.264282 & 0.128594 \\
\hline
per task adam8p (100 Trial) & 0.095372 & 0.046926 & 0.107177 & 0.046672 \\
\hline
per task adam8p (1k Trial) & 0.044144 & 0.028692 & 0.060663 & 0.028813 \\
\hline
per task adam8p + opt\_list(2k Trial) & 0.018403 & 0.023350 & 0.025419 & 0.025555 \\
\hline
\hline
\end{tabular}
\end{center}

\section{Experimental details for figures}
\subsection{Outer training of different learned architectures}
All models are trained with truncated evolutionary strategies using the parameters described in the main text.
Instead of using the two stage schedule we only train with truncation's of size 240-360.
We use the same, fixed learning rate for the FF and LSTM\_FF model of 0.000500. For LSTM we found a lower learning rate of 0.000100 to perform better.
Ideally we would include an extensive hyperparameter comparison, but the computational cost is prohibitive.

For the FF model we make use of a two hidden layer 32 unit MLP. In addition to the features described in \citep{metz2019understanding} we use RMS terms which improve performance over that reported in \citep{metz2019understanding} by a small amount.

For the L2LBGDGD model we copy \citet{andrychowicz2016learning} using a two hidden layer, 20 unit GRU~\citep{chung2014empirical} with the same input gradient processing. 

For the LSTM model, we make use of a model inspired by \citep{andrychowicz2016learning}.  We make use of a 64 dimension LSTM hidden size to produce two outputs, a direction and a sign which are combined similar to \citep{metz2019understanding}.

Performance on 100 tasks (each with five random seeds) is recorded over the course of training. We show an exponential moving average of this data in the figure.

\subsection{Outer training with different sized models}
All models are trained with truncated evolutionary strategies using parameters described in the main text.
Instead of using the two stage schedule we only train with truncation's of size 240-360.
We use a learning rate of $3*10^{-3}$ for all models.

\subsection{Large scale transfer with Resnet models}
We make use of two ResNetV2 models. Before applying the residual blocks, we pass the inputs through a 64 unit, 7x7 convolutional kernel with stride 2 followed by max pooling of size 3 with a stride of 2. This aggressive down sampling was originally designed for larger image models but was also left in despite our small sized models.

We then apply some number of residual blocks parameterized by a (number of output channels, number of bottleneck channels, stride).

At the output we apply batch norm, reduce over the spatial dimensions, and project to the the appropriate number of classes.

Our goal with these experiments are not to seek peak performance. Instead show that our learned optimizer is capable of generalizing to vastly different distributions of models.

\paragraph{CIFAR-10 Resnet}
This model makes use of four residual blocks: [(128, 32, 1), (128, 32, 2), (128, 64, 1), (128, 64, 2)].

\paragraph{Imagenet Resnet}
This model makes use of 11 residual blocks: [(128, 32, 1), (128, 32, 2), (256, 64, 1)*2, (256, 64, 2), (512, 128, 1)*3, (512, 128, 2), (1024, 256, 1)*2].

\subsection{Self optimization details}
For our baseline learned optimizer trained with Adam, we search over two learning rates, $10^{-5}$ and $3*10^{-3}$ and select the best one (1e-3) for the given 10k outer-training steps as well as selected the optimizer used to train the best optimizer (3e-5).

When training an optimizer with a new optimizer there are no hyper parameters. We found we must also include the original gradient clipping used for the Adam models (clipping all values between -0.1, 0.1). Without this, our learned optimizer is not able to make any progress. Understanding this is an interesting to us as we would have suspected the dynamic inner-clipping should have addressed this.

For technical reasons, the learned optimizer does not have easy access validation losses. Instead, we use the training loss for both. This is computed as the mean over the normalized losses computed over a single batch.

\section{Self Optimization Extended}
\label{app:selfopt}

\begin{figure}[h]
    \centering
    \includegraphics[width=0.45\textwidth]{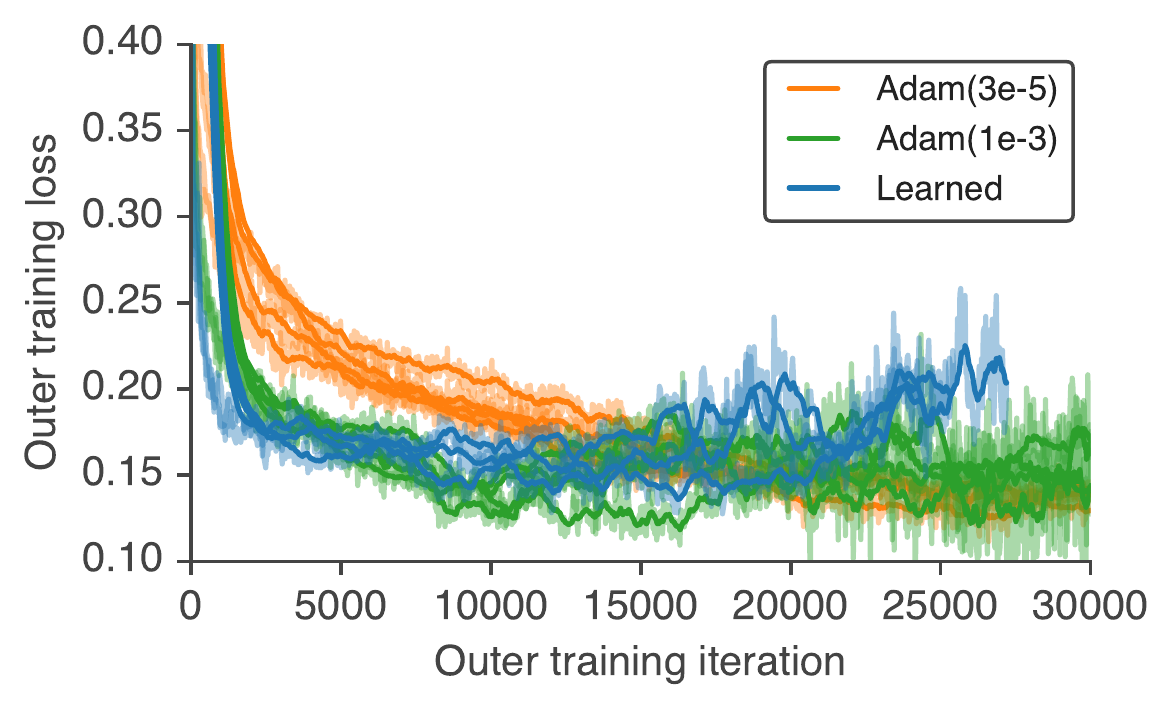}
    \caption{When using a learned optimizer to train itself, we find good performance early in training, but diverge after 10k outer-iterations. On the x-axis we show number of weight updates done to the learned optimizer. On the y-axis we show outer-loss. Each point consists of inner-training 100 models, each with five random initializations trained for 10k inner-iterations. We show the average validation performance post-normalization averaged over all tasks, seeds, and inner-training steps. Each line represents a different randomly initialized learned optimizer. In orange we show Adam with a learning rate of 3e-5 the same value used to train the optimizers in this work. In green, we show Adam with a learning rate of 1e-3, the learning rate that performed best in this 10k outer-iteration regime. In blue we show our learned optimizer. Light colors represent instantaneous outer-losses while solid denotes smoothing done via an exponential moving average.
    }
    \label{fig:self_opt}
\end{figure}

\clearpage

\section{Infrastructure Used}
\label{app:infra}
The types of learning systems we discuss here are more complex than the average deep learning training setup. We are training not one, but many different kinds of models on different datasets with wildly different properties. As a result, the static graph paradigm of traditional deep learning models -- distributed one graph and using MPI / allreduce style primitives -- is insufficient. Finally, evaluating the performance of an optimizer is not a trivial task as it requires inner-training a large number of different models for a large number of inner-iterations.

Developing custom training and evaluation infrastructure has been critical to the success of this family of work. We describe the training infrastructure employed, discuss techniques we use for evaluation and monitoring. At this time, we are unable to release a fully running open source implementation due to the use of internal tools.

\subsection{Outer-training cluster}
The cluster consists of six kinds of jobs -- workers, a learner, gradient storage, summary aggregators, evaluation chief, evaluation worker. Additionally we make use of a distributed file system \citep{ghemawat2003google}.

\begin{figure}[h]
    \centering
    \includegraphics[width=\textwidth]{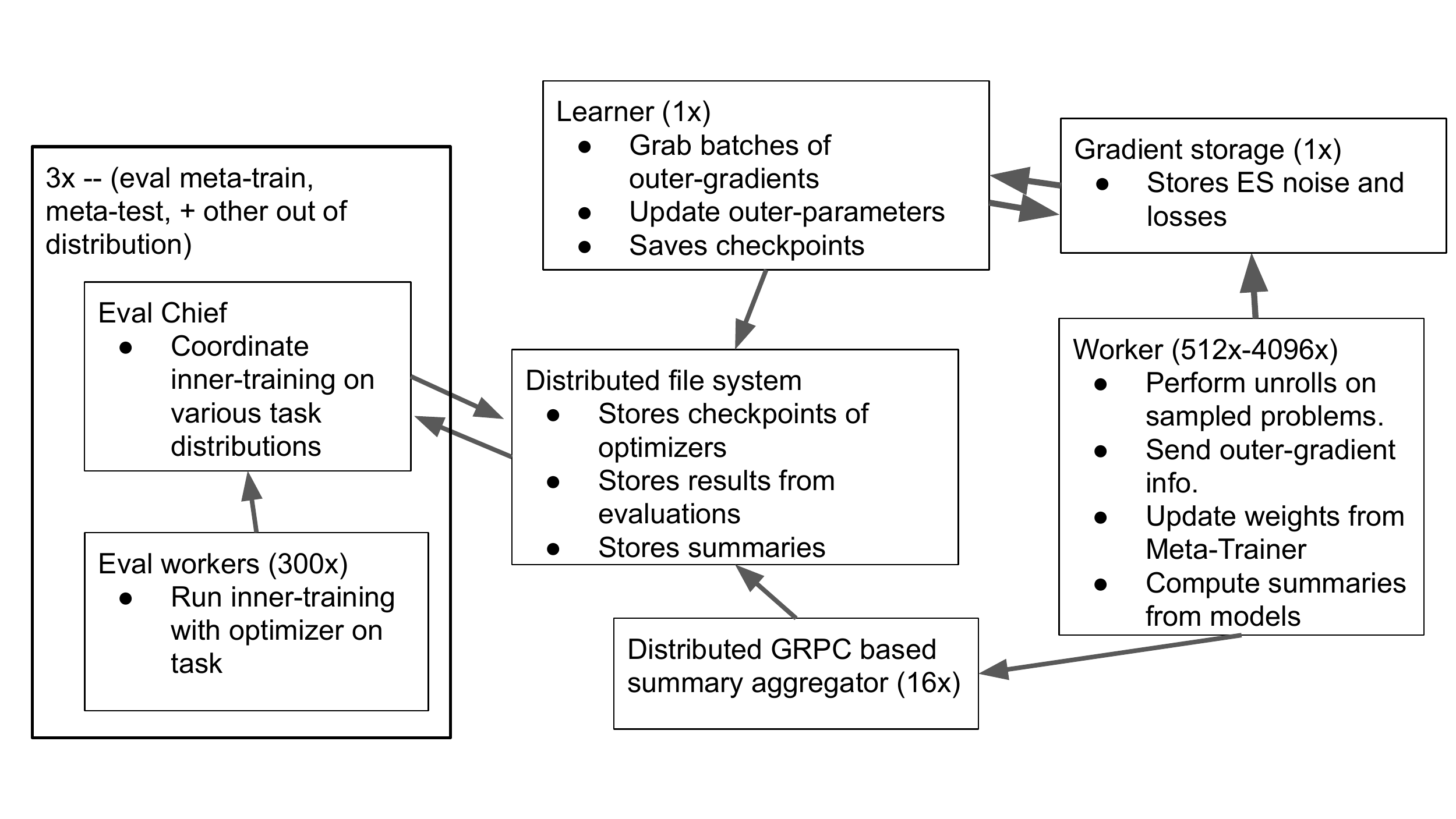}
    \caption{High level / system diagram for infrastructure used. See Appendix \ref{app:infra} for more information.
    \label{fig:infra}
    }
\end{figure}

\paragraph{Workers}
Workers are jobs that sample a outer-task, build a tensorflow \citep{abadi2016tensorflow} graph and provides some way to estimate outer-gradients.
In this work, we leverage antithetic ES sampling with shared randomness where possible.
In practice, this means we start two inner-training procedures starting from the same initializations, using the same minibatches of data, but leveraging two slightly differnt outer-parameter values.
Before each outer-gradient estimate new outer-parameters, as well as the current outer-iteration, are obtained by RPC call to the learner.
This gradient estimate, the current outer-iteration (e.g. global step) along with the type of problem selected is sent to the key value storage worker keyed with some unique identifier.

\paragraph{Gradient Storage}
The gradient storage job is a distributed dictionary which stores information computed from the worker.
At this point, given the low frequency of updates only a single machine is required.
This is a separate machine from the learner to control the number of queries required sent to the learner.

\paragraph{Learner}
The learner polls the gradient storage for batches of data.
If the outer-training iteration is within some fixed amount of time to the current outer iteration the data is deleted.
In this work we use 5.
This prevents stale gradients from being used to update the outer parameters.
This has the negative effect of potentially ignoring the slowest tasks. At this point, however, we find that this rejection threshold is quite small (<1\%) and thus safely ignored.
When a batch of outer-information is obtained, the learner aggregates outer-gradients, and performs an update of outer-optimization, in our case Adam.
Current loss values for each task family are also recorded in a dictionary, and aggregated across all tasks seen thus far. We find this averaged signal, as well as simply just logging out the loss from the current batch, to be useful for evaluating model performance as it is lower variance and is stationary in time.
This job is also responsible for writing the meta-parameters to disk every 10 min for use later and for evaluation.

\paragraph{Summary aggregators}
When training a standard model, users monitor values -- e.g. activation values, losses, or accuracies. In this work we are training thousands of different neural network tasks at a time and would ideally like to be able to obtain more insight into what is going on. Existing tooling such as TensorFlow's event files are inadequate. 

We work around this by introducing a distributed service meant just for summary aggregation. This service exposes a RPC api that takes batches of summaries and logs them to disk in a custom binary format for fast visualization later.
We shard request based on the name of the summary. We additionally log out subsampled data for ease of viewing after training.

In this work, we process roughly 10k summaries a second split across 24 machines. We found being able to do in-depth post-hoc analysis of model training to be critical to diagnosing issues and uncovering bugs. For example, looking at gradient variance for each task in the outer-training set let us diagnose which tasks are diverging and why.
To ensure that summary computation does not cause increased computational overhead we make use of both stochastic logging as well as both batching of RPC calls to summary aggregators, and sending these RPC in parallel with computation.

\paragraph{Evaluations:}
Evaluating a learned optimizer is not nearly as simple as say performing forward passes on some test set of minibatches. Evaluation consists of an expectation over training many different tasks. To obtain a low variance estimate we employ a large amount of compute to evaluate the checkpoints saved out by the learner. To lower variance, we pick a fixed set of tasks to evaluate. In addition to the meta-training task distribution (which we pick a subset of 100 to evaluate), we also evaluate on a test set of tasks which is iid to the meta-train but never seen when outer-training as well as an extremely out of distribution set of data not close to tasks in the outer-training distribution. By monitoring a variety of different problems, with different similarities to the meta-training set we obtain a better sense of outer-generalization. These evaluations take time, around an hour of compute per task, so it must parallelized.

Originally, we tried to avoid this complexity and obtain some measure of outer-loss directly from the training cluster. While this works to some extent, it is noisy, and doesn't capture what we actually care about -- performance of some fixed weights of a learned optimize when applied to a target problem for an extended period of time. Averaging over the meta-training cluster also doesn't capture variance from outer-weight to outer-weight. 

The evaluation of one task also requires orchestration. We employ a evaluation chief, and evaluation workers for each task set. While its possible to do this type of evaluation offline, we find running online yields a much better workflow and thus speeds up research.

\paragraph{Evaluation Chief:} The evaluation chief is responsible for watching directories on a distributed filesystem and to enqueue evaluation tasks. Workers use RPC to request, and to send back results (e.g. learning curves). When a set of tasks is complete for a given outer-parameter checkpoint the results are written out to disk for later analysis.

\paragraph{Evaluation workers:}
Evaluation workers request task configurations from the chief over RPC. This configuration is parsed and converted to a TensorFlow graph which is then run. Over the course of running, various signals including training loss, validation loss, and test loss, are recorded. Upon completion these results are sent back over RPC to the chief to be aggregated and written to disk.

\subsection{Computational expense}

At this point, training of a learned optimizer is quite expensive. It roughly entails 60K CPU cores for around a month, or around ~5k CPU years. Much as neural architecture search has been dramatically optimized and improved, we hope learned optimizers will obtain a similar outer-training speedups. The energy and environmental impact of this work is also worth noting. These models take on the order of 200 megawatt hours of power to outer-train. We believe that these methods can be used to speed up development of many existing models once outer-trained. By meta-training a hyper parameter free model, one has to do no hyper parameter tuning of the optimizer for example.  In general, one should weight the costs of outer-training against the potential savings achieved. In the future we expect these models will be trained once and applied everywhere.

Finally, we would like to highlight that all of this work was performed on CPU.
In tests on GPU (without optimizing) we found performance to be similar.
Modern accelerator hardware, TPU, GPU, are ill-suited for the small workloads we perform here (serially training lots of small networks).
We believe it is possible to vectorize the inner-training of neural networks (e.g. training N networks instead of just 1 per worker) but have not explored this route yet.

\end{document}